\documentclass{article}

\usepackage[nonatbib,preprint]{arxiv}

\usepackage[utf8]{inputenc}
\usepackage[ruled]{algorithm2e}
\usepackage{algcompatible}
\usepackage[intlimits]{amsmath}
\usepackage{titlesec}
\usepackage{float}
\usepackage{array}
\usepackage{multirow}
\usepackage[ngerman,english]{babel}
\usepackage{graphicx,wrapfig}
\usepackage[toc]{appendix}
\usepackage{etaremune}
\usepackage[inline]{enumitem}
\usepackage[onehalfspacing]{setspace}
\usepackage[T1]{fontenc}
\usepackage{amsfonts}
\usepackage{amsthm,amssymb}
\usepackage{mathtools}
\usepackage[most]{tcolorbox}
\usepackage{upgreek}
\usepackage[format=hang,font=small]{caption}
\usepackage[colorlinks,linkcolor=black,citecolor=black,urlcolor=black,breaklinks]{hyperref}
\usepackage{url}
\usepackage{booktabs}
\usepackage{nicefrac}
\usepackage{microtype}
\usepackage{fancyhdr}
\usepackage{cleveref}
\usepackage{adjustbox}
\usepackage{perpage}
\usepackage{doi}
\graphicspath{{Figs/}}
\MakePerPage{footnote}
\DeclareMathOperator*{\argmin}{arg\,min}
\DeclareMathOperator*{\argmax}{arg\,max}
\titlespacing{\section}{0pt}{5pt}{5pt}
\titlespacing{\subsection}{0pt}{\parskip}{-\parskip}
\titlespacing{\subsubsection}{0pt}{\parskip}{-\parskip}
\title{Hierarchical Quadratic Random Forest Classifier}
\author{Faezeh Fallah\\Institute of Signal Processing and System Theory\\University of Stuttgart,~Pfaffenwaldring 47\\70569 Stuttgart,~Germany\\\texttt{faezeh.fallah@iss.uni-stuttgart.de}}
\begin{document}
\maketitle
\begin{abstract}
In this paper, we proposed a hierarchical quadratic random forest classifier for classifying multiresolution samples extracted from multichannel data. This forest incorporated a penalized multivariate linear discriminant in each of its decision nodes and processed squared features to realize quadratic decision boundaries in the original feature space. The penalized discriminant was based on a multiclass sparse discriminant analysis and the penalization was based on a group Lasso regularizer which was an intermediate between the Lasso and the ridge regularizer. The classification probabilities estimated by this forest and the features learned by its decision nodes could be used standalone or foster graph-based classifiers.
\end{abstract}
\section{Background and Motivation}
\label{sec:BackMotivRandForest}
A famous problem of supervised learning in classification and regression tasks was the \textbf{bias-variance tradeoff}. It implied that a reduction in the estimation bias, i.e. enhancing the estimation accuracy, would compromise the estimation precision, i.e. increasing the estimation variance or reducing the generalization capability (predictive performance on unseen samples). This stemmed from the fact that a classifier of low bias had indeed fit well to its training samples and thus lost the opportunity to see the variations beyond these samples, i.e. what could happen in unseen (validation or test) samples.

To tackle this issue, ensemble classifiers got proposed. An \textbf{ensemble classifier} was composed of several \textbf{weak classifiers}. This allowed it to minimize its classification bias \textbf{within} each weak classifier while minimizing the classification variance \textbf{across} different weak classifiers. To minimize the classification bias or to maximize the classification accuracy on the training samples, every weak classifier should be a suboptimal (non-maximum-margin) classifier fitting well to its training samples, i.e. its decision boundaries should change with any small change in the training samples. To minimize the classification variance (overfitting or prediction error) or to enhance the generalization on unseen samples, different weak classifiers should be dissimilar to (independent from) each other. This could be achieved by feeding different training samples and/or features to different weak classifiers\label{indWeakClass}.

Every weak classifier was composed of one or more \textbf{basic classifier(s)}. A basic classifier could be of any kind of classifiers. However, a decision tree was a favorite one because it
\begin{itemize}[leftmargin=*]
\item was easy to implement and could process both continuous and categorical features without needing any normalization, scaling, or preprocessing;
\item was transparent (easily visible) by being composed of a root node at its top, some decision nodes in its middle, and some leaf nodes at its bottom;
\item could employ any discriminative classifier in each of its decision nodes.
\end{itemize}

The order of the minimization of the classification bias and variance of an ensemble classifier led to a \textbf{bagging} or a \textbf{boosting} scheme. The bagging scheme first minimized the classification variance by diversifying the weak classifiers and then minimized the classification bias by fitting each weak classifier to its training samples. The diversification of the weak classifiers was achieved by feeding a randomly selected subset of the training samples and features to each of them and optimizing them independent from (in parallel to) each other\label{bagging}. The boosting scheme first minimized the classification bias of each weak classifier by feeding all the training samples and features to it. Then, it minimized the classification variance by optimizing the parameters of the weak classifiers serially (one after another). That is, the optimization of each weak classifier aimed to reduce the classification variance already achieved by the previously optimized weak classifiers. To this end, at the end of the optimization of each weak classifier, its classification performance was measured on the training samples and a weight, reflecting the sample's difficulty for the classifier, got assigned to each training sample. Then, the training samples accompanied by their weights and features got used to optimize the next weak classifier. This classifier considered the samples' weights to focus on more difficult samples for a further reduction of the classification variance.

At the end of the optimization with the bagging or boosting, the classification performance of each decision/leaf node of the ensemble (forest) determined its contribution (weight) in classifying unseen samples. That is, each unseen sample got classified by all the weak classifiers. Then, a weighted average of those estimates defined the final estimete.

The boosting outperformed the bagging in dealing with missing features in some samples. However, its complexity, sensitivity to parameter variations, and tendency to overfitting were higher than the bagging. Accordingly, \textbf{the present work used the bagging}\label{bagging-boosting}. An analysis and comparison of different bagging and boosting approaches were in \cite{Galar2012}.
An ensemble classifier whose basic classifiers were decision trees and was collected according to a bagging scheme was called a random forest classifier. The term \textbf{random} referred to the randomly selected training samples and features for each weak classifier \cite{Breiman2001}. We assumed that this forest was composed of $n_{\mathrm{weak}}$ weak classifiers and classified samples into $n_{\mathrm{clas}}$ classes. The samples got fed to the root nodes of the decision trees of the weak classifiers of the forest. Each decision node classified its received samples into $n_{\mathrm{clas}}$ classes and then, according to these classifications, sent them to the nodes in the next (deeper) tree layer. This process continued until every sample reached a leaf node of a tree.

In the original random forest classifier, each decision node employed a univariate linear discriminant to classify its received samples. This discriminant found an optimum decision threshold for each feature element of the samples of each class separately and independently from other features. This allowed a \textbf{simultaneous multiclass classification}\footnote{We differentiated the \textbf{multiclass} from the \textbf{multilabel} classification. In both cases, the classifications were simultaneous not groupwise. However, in the former/latter each sample had one/more reference label(s).} but also resulted in axis-oriented decision boundaries in the feature space or an \textbf{orthogonal} random forest classifier \cite{Hastie1995,Breiman2001}. Such boundaries become nested and complex if the features were correlated or the classes overlapped. The nested and complex boundaries were prone to overfitting (small classification bias on the training samples and large classification variance on unseen samples). To mitigate this issue, a \textbf{multivariate (Fisher's) linear discriminant} got used in each decision node. This discriminant considered all the features together and reduced their dimensionality by linearly projecting them onto a line. The projection was done by multiplying the features with \textbf{optimum projecting coefficients} of the same dimension. Then, an \textbf{optimum decision threshold} was sought over the projection line to separate different classes. This resulted in a simple decision boundary in the form of an arbitrarily oriented affine set in the feature space or an \textbf{oblique} random forest classifier \cite{Menze2011}.

\begin{figure}[t!]
\begin{center}
\includegraphics[width=1.0\textwidth]{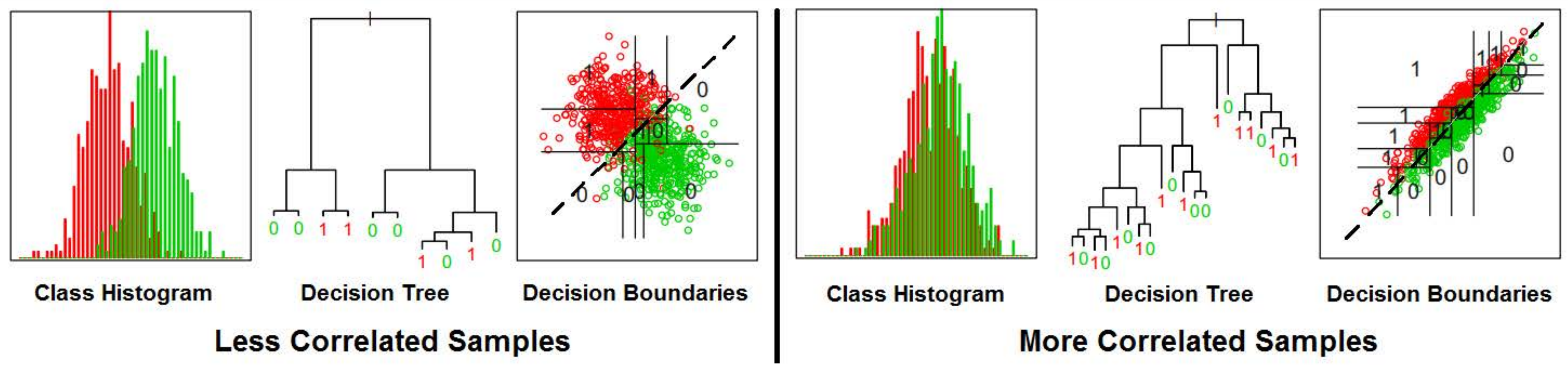}
\caption{The nested decision boundaries (solid lines) resulted from an application of a univariate linear discriminant and the simple decision boundaries (dashed lines) resulted from an application of a multivariate linear discriminant in the decision nodes of a decision tree.}
\label{fig:DecBndrUldMld}
\end{center}
\end{figure}

\autoref{fig:DecBndrUldMld} compares the decision boundaries of a univariate and a multivariate linear discriminant applied to the decision nodes of a decision tree. The combined features projection and the simple boundaries of the multivariate linear discriminant could mitigate features correlations and class overlaps and could reduce the chance of overfitting more than its univariate counterpart. However, it was \textbf{only applicable to a binary classification}.

We assumed that each sample $v_{r,j}$ of resolution $r\in\{1,\cdots,n_{\mathrm{lay}}\}$ had a reference classification label $l_{r,j}\in\mathbb{L}=\{1,\cdots,n_{\mathrm{clas}}\}$ and a feature vector $\mathbf{f}_{r,j}={[f_{r,j,h}]}_h\in\mathbb{R}^{n_{\mathrm{feat}}}$. We denoted a decision node at the $r^{\mathrm{th}}$ resolution layer of a decision tree of the $w^{\mathrm{th}}$ weak classifier of the forest with $\eta_{r,w}$. The set of the training samples received by this node was $\mathbb{D}_{r,w}$. The subset of this set containing the samples with a reference classification label of $c\in\mathbb{L}$ was $\mathbb{D}_{r,w,c}=\{v_{r,j}\in\mathbb{D}_{r,w}|l_{r,j}=c\in\mathbb{L}\}$ with $|\mathbb{D}_{r,w,c}|=n_{r,w,c}$. Then, we assumed that the samples in $\mathbb{D}_{r,w}$ were drawn from an $n_{\mathrm{feat}}$-variate Gaussian distribution. This distribution had a class-specific mean vector $\boldsymbol{\mu}_{r,w,c}\in\mathbb{R}^{n_{\mathrm{feat}}}$, a class-specific prior $\pi_{r,w,c}\in\mathbb{R}$, an overall mean vector $\boldsymbol{\mu}_{r,w}\in\mathbb{R}^{n_{\mathrm{feat}}}$, a common within-class covariance matrix $\boldsymbol{\Sigma}_{r,w}\in\mathbb{R}^{n_{\mathrm{feat}}\times n_{\mathrm{feat}}}$ for all the classes, and a between-class covariance matrix $\boldsymbol{B}_{r,w}\in\mathbb{R}^{n_{\mathrm{feat}}\times n_{\mathrm{feat}}}$ measuring the deviations of the class-specific means from the overall mean:
\begin{equation}
\label{eq:MLD_Params1}
\boldsymbol{\mu}_{r,w,c}=\frac{1}{n_{r,w,c}}\sum_{v_{r,j}\in\mathbb{D}_{r,w,c}}\mathbf{f}_{r,j}~~~~~~~~\pi_{r,w,c}=\frac{n_{r,w,c}}{\sum_{c'\in\mathbb{L}}n_{r,w,c'}}
\end{equation}
\begin{equation}
\label{eq:MLD_Params2}
\boldsymbol{\mu}_{r,w}=\sum_{c\in\mathbb{L}}\pi_{r,w,c}\cdot\boldsymbol{\mu}_{r,w,c}
\end{equation}
\begin{equation}
\label{eq:MLD_Params3}
\boldsymbol{\Sigma}_{r,w}=\frac{1}{\overbrace{(\sum_{c'\in\mathbb{L}}n_{r,w,c'})-n_{\mathrm{clas}}}^{\text{degree~of~freedom}}}\sum_{c\in\mathbb{L}}\sum_{v_{r,j}\in\mathbb{D}_{r,w,c}}(\mathbf{f}_{r,j}-\boldsymbol{\mu}_{r,w,c})\cdot(\mathbf{f}_{r,j}-\boldsymbol{\mu}_{r,w,c})^T
\end{equation}
\begin{equation}
\label{eq:MLD_Params4}
\boldsymbol{B}_{r,w}=\sum_{c\in\mathbb{L}}\pi_{r,w,c}\cdot(\boldsymbol{\mu}_{r,w,c}-\boldsymbol{\mu}_{r,w})\cdot(\boldsymbol{\mu}_{r,w,c}-\boldsymbol{\mu}_{r,w})^T.
\end{equation}

The inclusion of the degree of freedom in the denominator of $\boldsymbol{\Sigma}_{r,w}$ was to make it unbiased. The multivariate linear discriminant incorporated the Gaussian distributions of the samples into the Bayes' rule to derive their classification posteriors from their likelihoods and the priors of the samples and the classes. The prior distribution of the samples was independent from their class labels and thus was a constant factor. By considering the monotonic variation of the logarithms, the logarithms of the posteriors got taken. This turned the constant factor to a constant term. Then, based on a maximum a posteriori estimation, the objective function of the discriminant become the logarithm of the posteriors. With the constant term ignored, a maximization of this function returned the estimated label $\hat{l}_{r,j}\in\mathbb{L}$ for each sample $v_{r,j}$ as
\begin{subequations}
\label{eq:MLD}
\begin{align*}
\hat{l}_{r,j}=\argmax_{c\in\mathbb{L}}&\Big[\mathrm{log}(\pi_{r,w,c})+\boldsymbol{\beta}^T_{r,w,c}\cdot(\mathbf{f}_{r,j}-\frac{\boldsymbol{\mu}_{r,w,c}}{2})\Big]\\
\boldsymbol{\beta}_{r,w,c}&=\boldsymbol{\Sigma}_{r,w}^{-1}\cdot\boldsymbol{\mu}_{r,w,c}\tag{\ref{eq:MLD}}
\end{align*}
\end{subequations}
with $\boldsymbol{\beta}_{r,w,c}\in\mathbb{R}^{n_{\mathrm{feat}}}$ being the vector of optimum projecting coefficients for the class $c\in\mathbb{L}$ samples. Thus, the linear discriminant had a linear dependency on each $\mathbf{f}_{r,j}\in\mathbb{R}^{n_{\mathrm{feat}}}$ and needed to estimate $n_{\mathrm{feat}}$ parameters for each class. By adding the $n_{\mathrm{clas}}-1$ priors, its overall number of parameters become $n_{\mathrm{clas}}\cdot n_{\mathrm{feat}}+n_{\mathrm{clas}}-1$.

Also, by taking the class $c=1$ as the reference and defining the \textbf{discriminant directions} ${[\boldsymbol{\theta}_{r,w,c}\in\mathbb{R}^{n_{\mathrm{feat}}}]}_{c\in\mathbb{L}}$, \eqref{eq:MLD} turned into
\begin{equation}
\label{eq:MLD2}
\begin{split}
&\hat{l}_{r,j}=\argmax_{c\in\mathbb{L}}\Big[\mathrm{log}(\frac{\pi_{r,w,c}}{\pi_{r,w,1}})+\boldsymbol{\theta}^T_{r,w,c}\cdot(\mathbf{f}_{r,j}-\frac{\boldsymbol{\mu}_{r,w,1}+\boldsymbol{\mu}_{r,w,c}}{2})\Big]\\
&\boldsymbol{\theta}_{r,w,c}=\boldsymbol{\beta}_{r,w,c}-\boldsymbol{\beta}_{r,w,1}\implies\boldsymbol{\theta}_{r,w,1}=\mathbf{0}.
\end{split}
\end{equation}

\begin{figure}[t!]
\begin{center}
\includegraphics[width=0.7\textwidth]{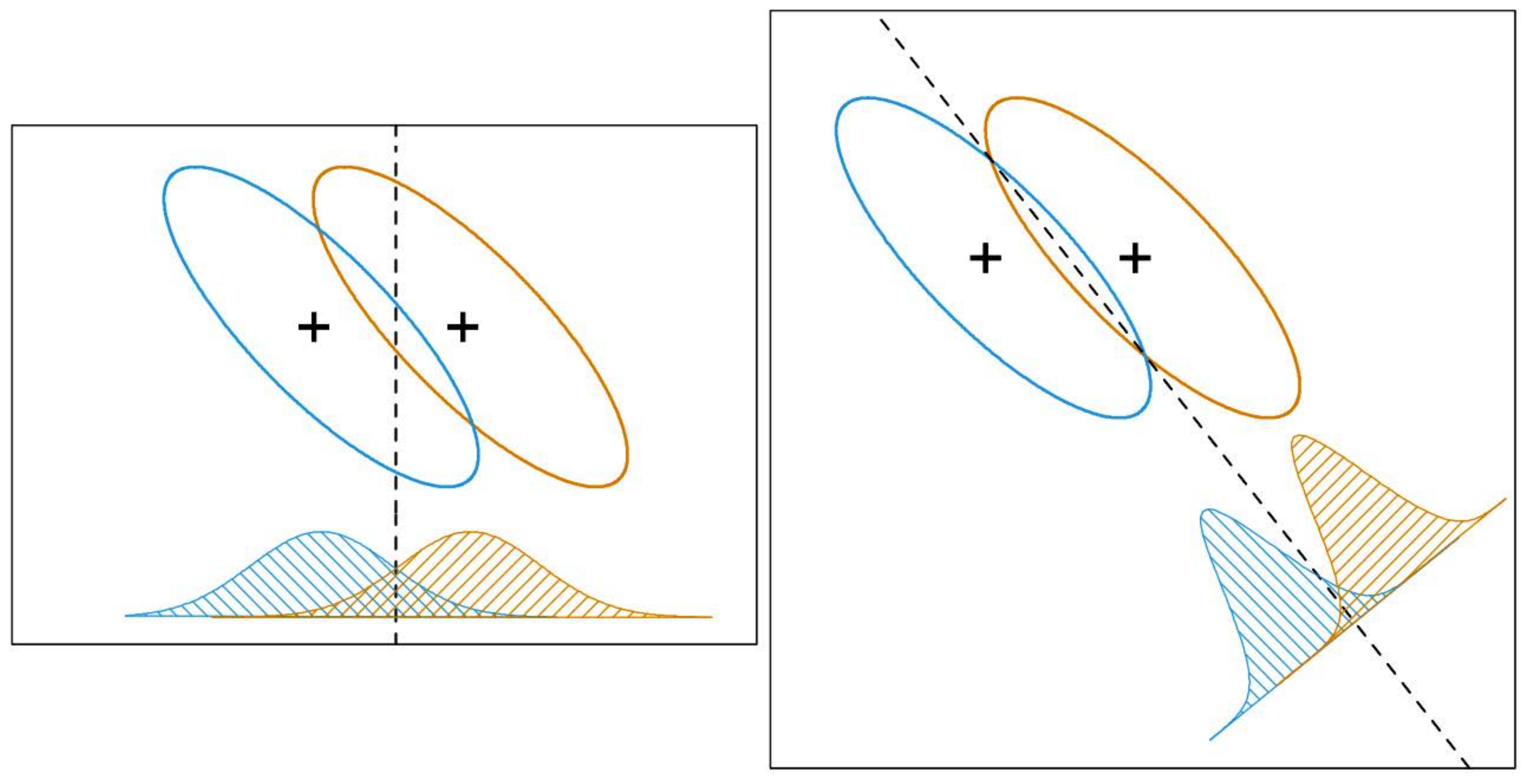}
\caption{a) Projection onto the line connecting the mean vectors of the classes maximized the distance between the means of the projected features but not the class separation. b) The class separation got maximized by projecting the features onto a line maximizing the Rayleigh quotient.}
\label{fig:RayleighQuotient}
\end{center}
\end{figure}

The above maximization was equivalent to the maximization of class separation (minimization of class overlap) in the lower dimensional space. It required to maximize the Rayleigh quotient (ratio of the between-class variance of all the projected features to the within-class variance of the projected features of each class). Then, by multiplying the optimum projecting coefficients with the features, these features got projected onto a line. As shown in \autoref{fig:RayleighQuotient}, this line was not necessarily the line connecting the mean vectors of the classes and thus did not necessarily maximized the distance between the means of the projected features. On this line, the optimum decision threshold was the point at which those classes were equally probable or had the same value of the objective function. The affine set perpendicular to the projection line and passing through this threshold was the decision boundary of the linear discriminant in the original feature space \cite{Yildiz2005,Hastie2017}.

The \textbf{multivariate quadratic discriminant} relaxed the requirement of having a common within-class covariance matrix for all the classes by allowing a class-specific within-class covariance matrix $\boldsymbol{\Sigma}_{r,w,c}\in\mathbb{R}^{n_{\mathrm{feat}}\times n_{\mathrm{feat}}}$ for each class $c\in\mathbb{L}$. This discriminant \textbf{was only applicable to a binary classification} and demanded the samples of each class to be from a multivariate Gaussian distribution. Based on these assumptions, the Bayes' rule, and the maximum a posterior estimation, the objective function of this discriminant got formed and its maximization returned the estimated label $\hat{l}_{r,j}\in\mathbb{L}$ for each sample $v_{r,j}$ as
\begin{equation}
\label{eq:QLD}
\hat{l}_{r,j}=\argmax_{c\in\mathbb{L}}\Big[\mathrm{log}(\pi_{r,w,c})-\frac{\mathrm{log}(|\boldsymbol{\Sigma}_{r,w,c}|)}{2}-(\mathbf{f}_{r,j}-\boldsymbol{\mu}_{r,w,c})^T\cdot\frac{\boldsymbol{\Sigma}^{-1}_{r,w,c}}{2}\cdot(\mathbf{f}_{r,j}-\boldsymbol{\mu}_{r,w,c})\Big]
\end{equation}
\begin{equation}
\label{eq:classCovMtrx}
\boldsymbol{\Sigma}_{r,w,c}=\frac{1}{n_{r,w,c}}\sum_{v_{r,j}\in\mathbb{D}_{r,w,c}}(\mathbf{f}_{r,j}-\boldsymbol{\mu}_{r,w,c})\cdot(\mathbf{f}_{r,j}-\boldsymbol{\mu}_{r,w,c})^T.
\end{equation}

\begin{figure}[t!]
\begin{center}
\includegraphics[width=1.0\textwidth]{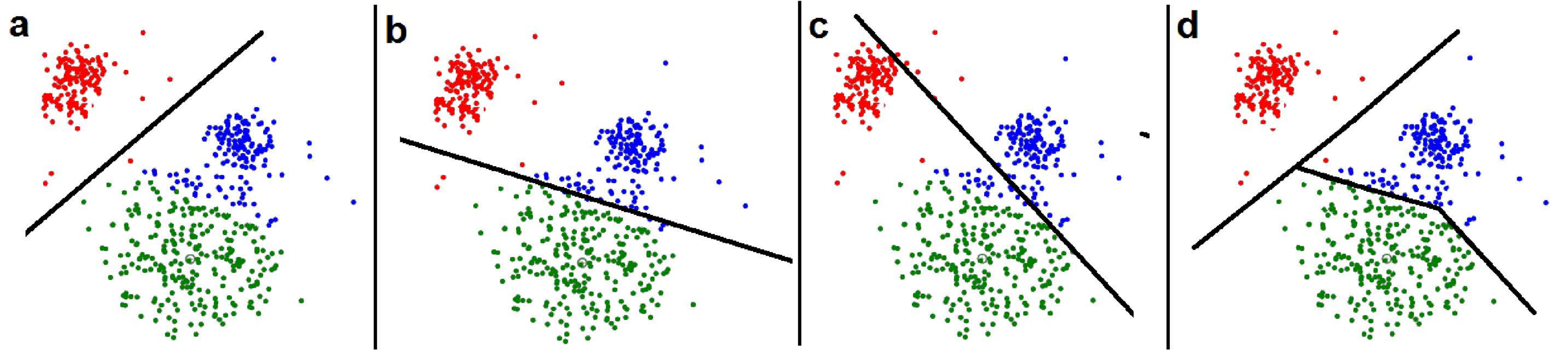}
\caption{a, b, c) Decision boundaries of 3 one-vs-all classifiers in a two-dimensional feature space. d) Final (combined) decision boundaries.}
\label{fig:OneVsAll}
\end{center}
\end{figure}

The decision boundary of this discriminant in the original feature space was quadratic and indeed a conic section (i.e. a line, a circle, a parabola, an ellipse, or a hyperbola). In comparison to the affine set boundary of the linear discriminant, this quadratic boundary could capture more complex class separations and could thereby reduce the classification bias of the decision tree using it in its decision nodes. However, this higher flexibility come at the cost of a higher number of parameters to estimate. The quadratic discriminant had a quadratic dependency on each feature vector $\mathbf{f}_{r,j}$ and thus needed to estimate $n_{\mathrm{feat}}\cdot(n_{\mathrm{feat}}+1)/2$ parameters for each class-specific covariance matrix\footnote{A covariance matrix was always symmetric positive semi-definite.}. By adding the $n_{\mathrm{clas}}-1$ priors, its overall number of parameters become $n_{\mathrm{clas}}\cdot n_{\mathrm{feat}}\cdot(n_{\mathrm{feat}}+1)/2+n_{\mathrm{clas}}-1$ \cite{Hastie2017}.

Moreover, both the linear and the quadratic discriminant were only applicable to binary classifications. To apply them to a classification on $n_{\mathrm{clas}}>2\label{one-vs-all}$ classes, the multiclass classification should be decomposed into a set of binary classifications and then the results of these classifications should be combined. The set could be the result of $\binom{n_{\mathrm{clas}}}{2}$ combinatorial selections or $n_{\mathrm{clas}}$ one-vs-all selections or a hierarchical superclass-to-microclass classification. In the first two approaches, a major vote should be collected across the results of the binary classifications to find the final estimate. \autoref{fig:OneVsAll} shows a 3-class classification through 3 one-vs-all classifications. In the third approach, the classifications were done over multiple steps. In the first step, the binary classification got conducted on two superclasses formed from the $n_{\mathrm{clas}}$ classes. In the next step, the more important superclass got decomposed into two finer superclasses and the binary classification got conducted on these new classes. These steps got continued until the finest class (microclass) was reached. In comparison to a simultaneous multiclass classification, each of these approaches had a higher complexity and a higher memory footprint. In addition, the superclass-to-microclass classification could propagate errors from a coarser classification level to a finer classification level.

\begin{figure}[t!]
\begin{center}
\includegraphics[width=0.9\textwidth]{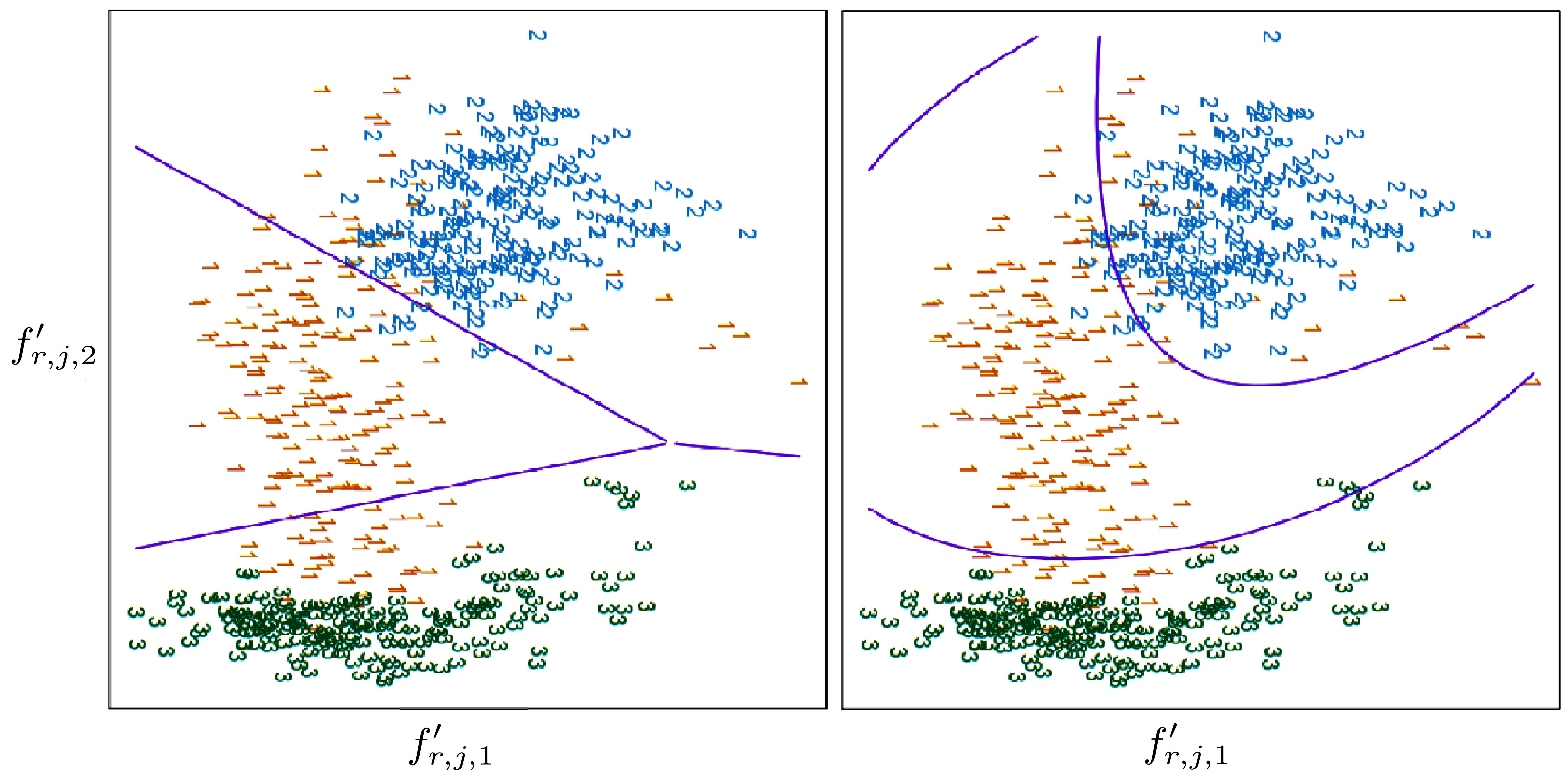}
\caption{\textbf{Left}: Linear boundaries of a linear discriminant applied to the original features, i.e. $\mathbf{f}_{r,j}=[f_{r,j,1},f_{r,j,2}]=[f'_{r,j,1},f'_{r,j,2}]$. \textbf{Right}: Quadratic boundaries in the original feature space after applying the linear discriminant to the squared features, i.e. $\mathbf{f}_{r,j}=[f_{r,j,1},\cdots,f_{r,j,5}]=[f'_{r,j,1},f'_{r,j,2},(f'_{r,j,1}\cdot f'_{r,j,2}),(f'^2_{r,j,1}),(f'^2_{r,j,2})]$.}
\label{fig:LinearQuadraticBoundary}
\end{center}
\end{figure}

\textbf{The linear discriminant could realize the quadratic decision boundaries and thus the lower classification bias of the quadratic discriminant if it was applied to squared features}. This was due to the fact that the linear boundaries of the linear discriminant in the squared feature space translated to quadratic boundaries in the original (unsquared) feature space. \autoref{fig:LinearQuadraticBoundary} shows this concept for two features and three classes.

As given by \eqref{eq:MLD} and \eqref{eq:QLD}, both the linear and the quadratic discriminant needed to compute inverses of the covariance matrices to estimate the classification labels of the samples. Thus, they required the covariance matrices to be \textbf{full rank} or \textbf{invertible}. The rank deficiency of the covariance matrix $\boldsymbol{\Sigma}_{r,w,c}\in\mathbb{R}^{n_{\mathrm{feat}}\times n_{\mathrm{feat}}}$ of a class $c\in\mathbb{L}$ could arise from
\begin{enumerate}[label={(\arabic*)},leftmargin=*]
\item having more features than the number of training samples for the class $c\in\mathbb{L}$.
\item linear dependency of some columns of $\boldsymbol{\Sigma}_{r,w,c}$ due to the correlation of the corresponding feature vectors \footnote{Independence implied being uncorrelated but the converse was not always true. Assuming that the features were drawn from a multivariate Gaussian distribution made the correlation and the dependence equivalent.}, i.e. one vector was a constant multiple of another vector.
\end{enumerate}

If the covariance matrix $\boldsymbol{\Sigma}_{r,w,c}$ was real- or complex-valued, then its noninvertibility could be tackled by computing its pseudoinverse through a least squares fit to its system of linear equations. This system had no unique solution. Thus, the fit was done by computing the singular value decomposition of $\boldsymbol{\Sigma}_{r,w,c}$. This decomposition could fail if $\boldsymbol{\Sigma}_{r,w,c}^T\cdot\boldsymbol{\Sigma}_{r,w,c}$ was singular. In this case, the least squares fit could not be found uniquely. In addition, any least squares estimator minimized its bias at the cost of increasing its variance. These made the pseudoinverse computation not a stable solution for many applications.

\begin{wrapfigure}{R}{8cm}
\centering
\includegraphics[width=8cm]{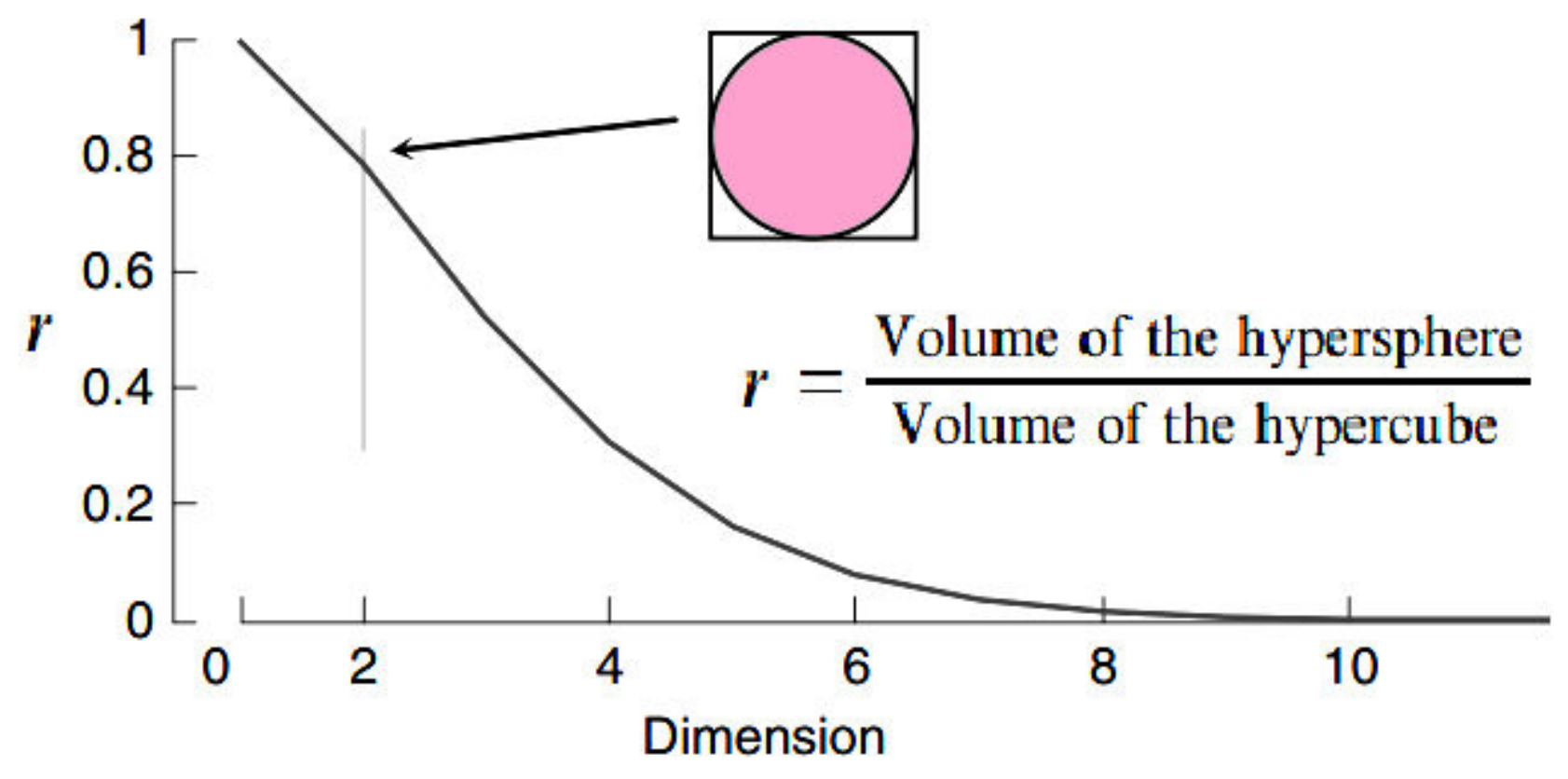}
\captionsetup{format=plain}
\caption{Drastical reduction of the proportion of the volume occupied by an inscribed hypersphere to the volume of its circumscribed hypercube after an increase in the dimension of the Euclidean space.}
\label{fig:CurseOfDimension}
\end{wrapfigure}

Moreover, any classifier acting in an Euclidean feature space could encounter with the \textbf{curse of dimensionality}. This occurred when an increase in the feature dimension reduced distances between the samples. As shown in \autoref{fig:CurseOfDimension}, when the feature dimension increased, the proportion of the volume occupied by an inscribed hypersphere to the volume of its circumscribed hypercube reduced drastically. Consequently, if $10^3$ evenly-spaced samples were enough to sample a one-dimensional unit interval with $0.001$ spacing between adjacent samples, then a sampling of a 10-dimensional unit hypercube with the same spacing between the adjacent samples would need ${(10^3)}^{10}=10^{30}$ samples. Thus, a factor $10$ increase in the dimension demanded a factor $10^{30}/10^3=10^{27}$ increase in the number of training samples. This way, the curse of dimensionality demanded an exponential rise in the number of training samples to still cover the new feature space sufficiently and thereby maintain the discriminative power of the classifier. Accordingly, this curse become more prominent when a limited number of training samples was available to optimize the classifier parameters with regard to a large feature dimension \cite{Keogh2010}. Another bottleneck of any projection-based classifier was computation and use of \textbf{large projecting coefficients}.

The large magnitudes of the coefficients reflected a high classification variance (overfitting) or a poor predictive performance (generalization) on unseen samples. These stemmed from existence of correlated features in the feature vectors of the training samples. Accordingly, it was desired to find and select \textbf{(uncorrelated) most discriminant features} for each classification task. In addition, to tackle the rank deficiency of the samples' covariance matrices, the curse of dimensionality (high-dimensional feature vectors), and the overfitting, shrinkage (compression) techniques were proposed. These techniques increased the classification bias slightly but were more numerically stable than the pseudoinverse computation. They got applied by penalizing\footnote{The terms \textit{penalization} and \textit{regularization} were used interchangeably.} or constraining the projecting coefficients \cite{Hastie2017}.

For the projecting coefficients $\boldsymbol{\beta}_{r,w,c}\in\mathbb{R}^{n_{\mathrm{feat}}}$ defined in \eqref{eq:MLD}, the penalization term was $\lambda_{r,\mathrm{lda}}\cdot{\|\boldsymbol{\beta}_{r,w,c}\|}_q^q$ with $\lambda_{r,\mathrm{lda}}\in\mathbb{R}_{+}$ being a regularization hyperparameter. This term got added to the sum of squared errors (differences between the estimated and the reference classification labels) to form the objective function of the discriminant like a penalized least squares estimator. Alternatively, the objective function could be the sum of squared errors constrained with ${\|\boldsymbol{\beta}_{r,w,c}\|}_q^q$ being less than or equal to a certain threshold. This resulted in a constrained least squares estimator. Thus, there was a one-to-one correspondence between the regularization hyperparameter $\lambda_{r,\mathrm{lda}}\in\mathbb{R}_{+}$ and the threshold. This implied that the optimum projecting coefficients were the solutions of the penalized least squares or the intersections of the contours (level sets) of the sum of squared errors and the constraint.

\begin{figure}[t!]
\begin{center}
\includegraphics[width=0.9\textwidth]{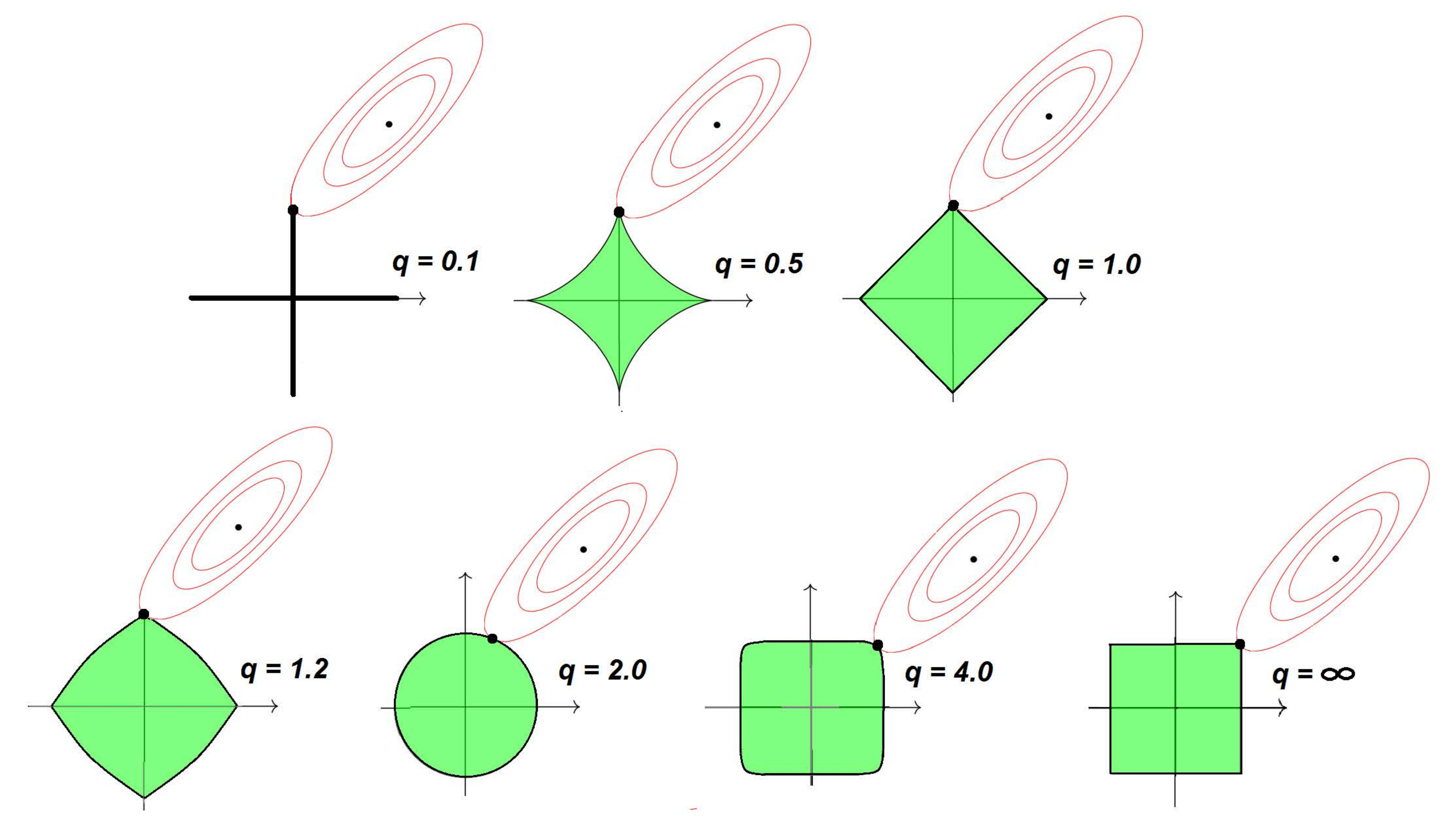}
\caption{Intersections between red contours of sum of squared errors and ${\|\boldsymbol{\beta}_{r,w,c}\|}_q^q$ for $\mathrm{dim}(\boldsymbol{\beta}_{r,w,c})=2$ and various values of $q$.}
\label{fig:PenLeastSqrVarNorm}
\end{center}
\end{figure}

\autoref{fig:PenLeastSqrVarNorm} shows these intersections and the contours of constant ${\|\boldsymbol{\beta}_{r,w,c}\|}_q^q$ for $\mathrm{dim}(\boldsymbol{\beta}_{r,w,c})=2$ and various values of $q$. Two special cases of the penalization resulted from the $l_1$ norm ($q=1$) and the $l_2$ (Frobenius) norm ($q=2$) of the projecting coefficients. The former led to a \textbf{Lasso}\footnote{Least absolute shrinkage and selection operator also called basis pursuit} and the latter led to a \textbf{ridge regularization}. As shown in \autoref{fig:PenLeastSqrVarNorm}, both of these regularizers were convex. However, in regards to a Bayes estimate, they represented different kinds of priors and thus led to different posteriors. The Lasso regularizer realized mode (maximum) of the posteriors and the ridge regularizer realized mean of the posteriors. Also, as depicted in \autoref{fig:PenLeastSqrVarNorm}, the Lasso regularizer with $q=1$ formed a diamond-shaped region and the ridge regularizer with $q=2$ formed a disk-shaped region in a two-dimensional feature space. Unlike the disk, the diamond had corners on the axes. If an intersection happened in one of these corners, then the corresponding projecting coefficient become zero. In dimensions higher than 2, the disk become a ball and the diamond become a rhomboid. The rhomboid had several edges and faces. An intersection in one of these edges or faces forced some of the projecting coefficients to zero. This way, the Lasso regularizer could \textbf{explicitly select most discriminant features} and could thereby \textbf{realize a sparse feature space}. In contrast, the round shape of the quadratic constraint of the ridge regularizer compressed each projecting coefficient in proportion to its squared magnitude and allowed the intersections to occur at more arbitrary places but not necessarily at the axes where some coefficients got vanished.

\begin{figure}[t!]
\begin{center}
\includegraphics[width=1.0\textwidth]{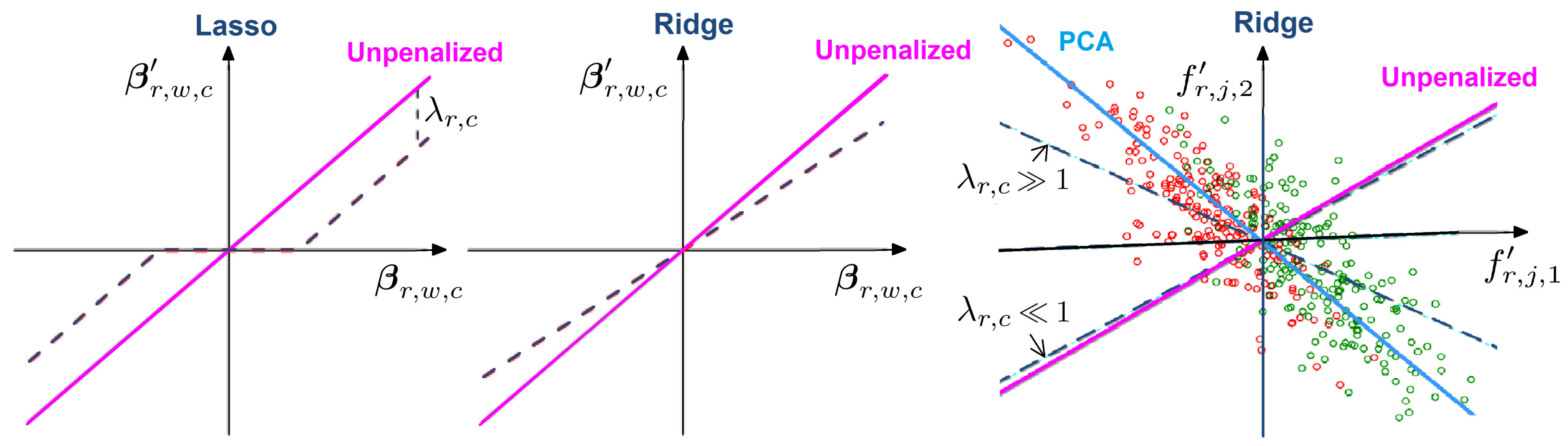}
\caption{The transformations of the projecting coefficients (depicted by dashed lines) after applying the Lasso or the ridge regularizer to a multivariate linear discriminant. The incorporation of the unsupervised principal component analysis (PCA) into the ridge regularizer and its effects (depicted by dashed lines) on the decision boundary of a multivariate linear discriminant in a two-dimensional feature space.}
\label{fig:LassoRidge}
\end{center}
\end{figure}

By considering a common within-class covariance matrix $\boldsymbol{\Sigma}_{r,w}\in\mathbb{R}^{n_{\mathrm{feat}}\times n_{\mathrm{feat}}}$ for all the classes and defining $\boldsymbol{\sigma}_{r,w}=\mathbf{diag}(\boldsymbol{\Sigma}_{r,w})$ to contain features variances, the inclusion of the Lasso or the ridge regularizer into the objective function of the multivariate linear discriminant resulted in a transformation of the optimum projecting coefficients $\boldsymbol{\beta}_{r,w,c}\in\mathbb{R}^{n_{\mathrm{feat}}}$ to
\begin{equation}
\label{eq:LassoRidgeTrans}
\begin{split}
\text{Lasso}:&~~~~\boldsymbol{\beta}'_{r,w,c}=\mathrm{sign}(\boldsymbol{\beta}_{r,w,c})\cdot\mathrm{ReLU}\big(|\boldsymbol{\beta}_{r,w,c}|-\lambda_{r,\mathrm{lda}}\big)\\
\text{Ridge}:&~~~~\boldsymbol{\beta}'_{r,w,c}=\boldsymbol{\beta}_{r,w,c}\odot\big(\boldsymbol{\sigma}_{r,w}\oslash(\boldsymbol{\sigma}_{r,w}+\lambda_{r,\mathrm{lda}}\cdot\mathbf{1})\big).
\end{split}
\end{equation}
That is, the Lasso regularizer translated each projecting coefficient by $\lambda_{r,\mathrm{lda}}\in\mathbb{R}_{+}$ and then truncated it at zero. This way, it could eliminate the singular (zero eigenvalue) dimensions of the covariance matrix $\boldsymbol{\Sigma}_{r,w}\in\mathbb{R}^{n_{\mathrm{feat}}\times n_{\mathrm{feat}}}$ and thus mitigate its rank-deficiency (singularity) and noninvertibility. The ridge regularizer mitigated the rank deficiency of the covariance matrix $\boldsymbol{\Sigma}_{r,w}\in\mathbb{R}^{n_{\mathrm{feat}}\times n_{\mathrm{feat}}}$ by transforming it to $\boldsymbol{\Sigma}'_{r,w}=(\boldsymbol{\Sigma}_{r,w}+\lambda_{r,\mathrm{lda}}\cdot\mathbf{I})$. Since every covariance matrix was symmetric positive definite and $\lambda_{r,\mathrm{lda}}\in\mathbb{R}_{+}$ was nonnegative, the ridge regularizer kept the optimizations convex and well-posed\footnote{Solution of a well-posed problem changed continuously with the continuous changes of its initial conditions.} and guaranteed the existence and uniqueness of the solution. \autoref{fig:LassoRidge} shows these transformations by comparing the unpenalized projecting coefficients $\boldsymbol{\beta}_{r,w,c}\in\mathbb{R}^{n_{\mathrm{feat}}}$ with the penalized ones $\boldsymbol{\beta}'_{r,w,c}\in\mathbb{R}^{n_{\mathrm{feat}}}$.

The \textbf{nonlinear transformation} of the Lasso regularizer was similar to the soft thresholding applied in wavelet-based smoothing. It resulted in a \textbf{nonlinearly penalized linear least squares estimator} whose optimal solutions (projecting coefficients) had no closed form and should thus be found through an iterative optimization. In contrast, the ridge regularizer \textbf{linearly transformed} the projecting coefficients by incorporating an unsupervised principal component analysis (PCA). The PCA found variance of the features of the samples of each class along the principal direction specified by every eigenvector of the class' covariance matrix. Accordingly, the ridge regularizer shrunk (compressed) each element of the projecting coefficients of each class with the measured variance of the class' features in the direction of that element. More specifically, if $\sigma_{i}^2$ was the variance of the features of a class' samples along the direction $i$, then the projecting coefficient for this direction got multiplied by $\sigma_{i}^2/(\sigma_{i}^2+\lambda_{r,\mathrm{lda}})$. Thus, the lower the variance $\sigma_{i}^2\geq 0$ was, the lower the $\sigma_{i}^2/(\sigma_{i}^2+\lambda_{r,\mathrm{lda}})\geq 0$ and the more the compression would be. This resulted in a \textbf{linearly penalized linear least squares estimator} whose optimal solutions (projecting coefficients) had a closed form and could be found by vanishing the gradients of its objective function with respect to the projecting coefficients. The ridge regularizer was also called weight decay \cite{Boyd2004,Hastie2017}.

The simplicity and the low parameter size of the multivariate linear discriminant made it an attractive choice for many applications. Accordingly, several extensions of this discriminant got proposed to tackle its bottlenecks. The semiparametric linear discriminant relaxed the requirement of having features of each class' samples from a multivariate Gaussian distribution. This was achieved by estimating a set of strictly monotone univariate transformations which could transform the features' distributions to the desired multivariate Gaussian distributions. These transformations got derived from the training samples and kept the Bayes' rule still linear with respect to the transformed features. Other extensions of the linear discriminant focused on the estimation of a sparse vector of projecting coefficients through the Bayes rule. This sparsity aimed to select a subset of features in order to tackle large features dimension or features dimension larger than the number of training samples. A toolbox called TULIP provided efficient implementations of these proposals \cite{Lin2003,Pan2021}.

\begin{figure}[t!]
\begin{center}
\includegraphics[width=0.9\textwidth]{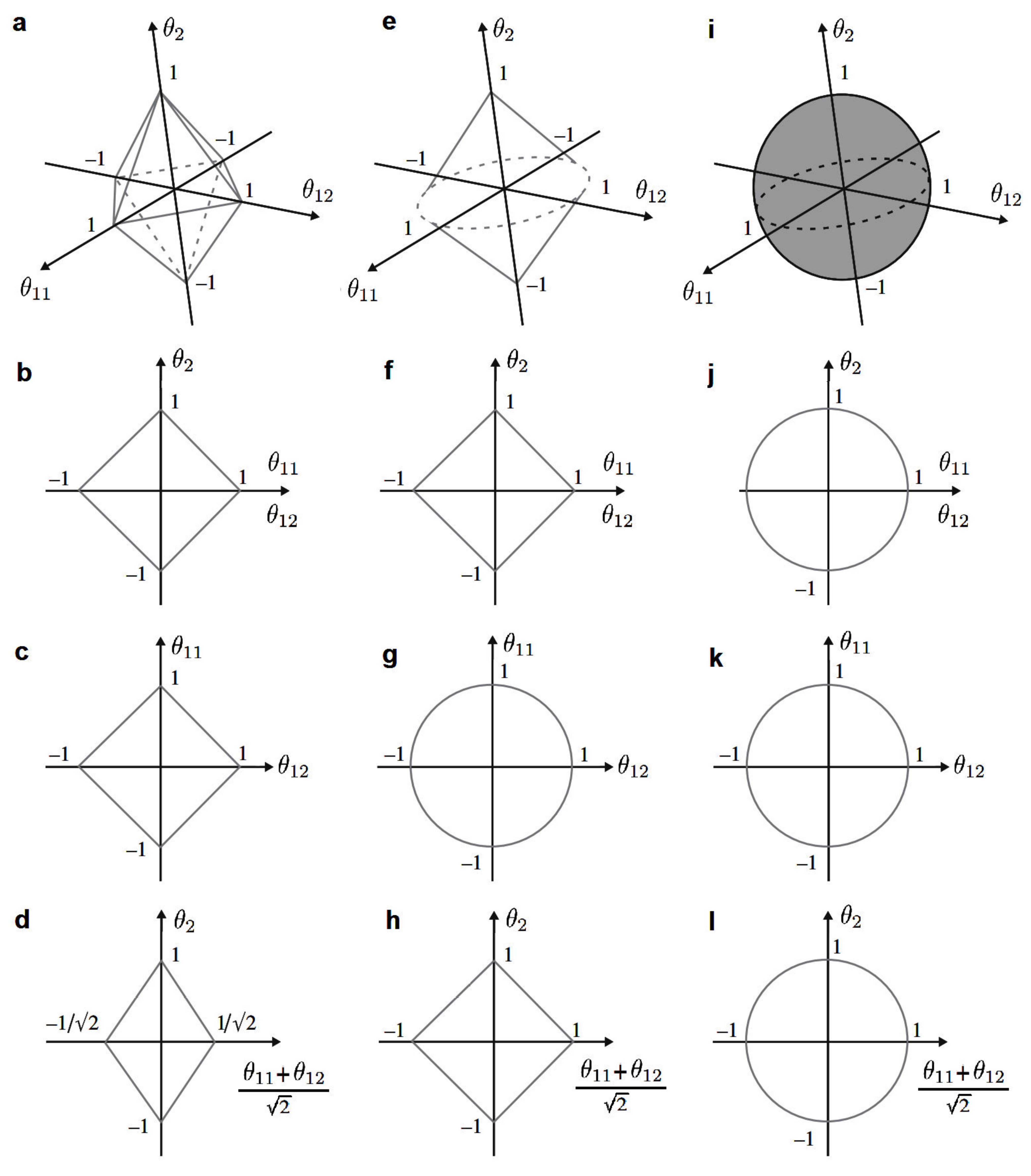}
\caption{a--d) Contour of ${\|\boldsymbol{\theta}\|}_1=1$ and its intersections with different planes. e--h) Contour of ${\|\boldsymbol{\theta}\|}_{\mathrm{gl}}=1$ and its intersections. i--l) Contour of ${\|\boldsymbol{\theta}\|}_2=1$ and its intersections.}
\label{fig:GroupLassoNorm}
\end{center}
\end{figure}

Each of these proposals, except for a multiclass sparse discriminant analysis (MSDA), used the Lasso regularizer to find its discriminant features. This implied that it computed its discriminant directions sequentially (one by one) and thus was limited to a binary classification. As described in page~\pageref{one-vs-all}, to apply such methods to a multiclass classification task, the task should be decomposed into a series of binary classifications. Then, for each binary classification the respective discriminant direction should be computed. This sequential computation was subject to error propagation and could not utilize the between-class information. Accordingly, its performance in feature selection was also limited. In contrast, the MSDA computed all the discriminant directions simultaneously and thus allowed a simultaneous multiclass classification. This was achieved by using a \textbf{group Lasso} regularizer which was an intermediate between the Lasso and the ridge regularizer.

The group Lasso regularizer selected features which influenced the Bayes rule. The influence of the $h^{\mathrm{th}}$ elements of the samples' features on the Bayes rule was reflected by the $h^{\mathrm{th}}$ elements of the discriminant directions ${[\boldsymbol{\theta}_{r,w,c}\in\mathbb{R}^{n_{\mathrm{feat}}}]}_{c\in\mathbb{L}}$ introduced in \eqref{eq:MLD2} \cite{Yuan2006,Mai2019}. Regarding $\boldsymbol{\theta}_{r,w,1}=\mathbf{0}$, this influence vanished if
\begin{equation}
\label{eq:disDirVanish}
\theta_{r,w,2,h}=\cdots=\theta_{r,w,n_{\mathrm{clas}},h}=0~~~~~~~~h\in\{1,\cdots,n_{\mathrm{feat}}\}.
\end{equation}
Accordingly, indices of the features selected by the group Lasso regularizer were indices of the discriminant directions which did not fulfill \eqref{eq:disDirVanish}. This implied that for each feature index $h$ the $h^{\mathrm{th}}$ elements of all the $n_{\mathrm{clas}}-1$ discriminant directions should be grouped together to be considered together. This group formed the vector
\begin{equation}
\label{eq:groupDisDir}
\boldsymbol{\theta}_{r,w,(2:n_{\mathrm{clas}}),h}=[\theta_{r,w,2,h},\cdots,\theta_{r,w,n_{\mathrm{clas}},h}]\in\mathbb{R}^{n_{\mathrm{clas}}-1}.
\end{equation}
The group Lasso regularizer selected the discriminant features by applying a Lasso regularizer \textbf{across} the groups of discriminant directions and a ridge regularizer \textbf{inside} each group. For a grouped vector like $\boldsymbol{\theta}=[\boldsymbol{\theta}_1,\theta_2]=\big[[\theta_{11},\theta_{12}],\theta_2\big]$, the $l_1$ norm used by the Lasso regularizer was ${\|\boldsymbol{\theta}\|}_1=|\theta_{11}|+|\theta_{12}|+|\theta_2|$ and the $l_2$ norm used by the ridge regularizer was ${\|\boldsymbol{\theta}\|}_2=\sqrt{\theta_{11}^2+\theta_{12}^2+\theta_2^2}$. However, the norm used by the group Lasso regularizer was ${\|\boldsymbol{\theta}\|}_{\mathrm{gl}}={\|\boldsymbol{\theta}_1\|}_{2}+|\theta_2|=\sqrt{\theta_{11}^2+\theta_{12}^2}+|\theta_2|$. \autoref{fig:GroupLassoNorm} shows contours of ${\|\boldsymbol{\theta}\|}_1=1$, ${\|\boldsymbol{\theta}\|}_{\mathrm{gl}}=1$, and ${\|\boldsymbol{\theta}\|}_2=1$ along with the intersections of these contours with different planes \cite{Yuan2006,Mai2019}.

\begin{figure}[!t]
\begin{center}
\fbox{\begin{minipage}{0.9\textwidth}
\textbf{Lemma~1:}~Given $\{\boldsymbol{\theta}_{r,w,(2:n_{\mathrm{clas}}),h'}|h'\neq h\}$ and defining
\begin{equation}
\label{eq:tildeTheta}
\begin{split}
&\boldsymbol{\delta}_c=\boldsymbol{\mu}_{r,w,c}-\boldsymbol{\mu}_{r,w,1}={[\delta_{c,h}]}_h\in\mathbb{R}^{n_{\mathrm{feat}}}\\
&\tilde{\theta}_{c,h}=(\delta_{c,h}-\sum_{h'\neq h}\theta_{r,w,c,h'}\cdot{[\boldsymbol{\Sigma}_{r,w}]}_{h',h})/{[\boldsymbol{\Sigma}_{r,w}]}_{h,h}\\
&\tilde{\boldsymbol{\theta}}_{(2:n_{\mathrm{clas}}),h}=[\tilde{\theta}_{2,h},\cdots,\tilde{\theta}_{n_{\mathrm{clas}},h}]\in\mathbb{R}^{n_{\mathrm{clas}}-1}
\end{split}
\end{equation}
allowed to express $\boldsymbol{\theta}_{r,w,(2:n_{\mathrm{clas}}),h}\in\mathbb{R}^{n_{\mathrm{clas}}-1}$ as
\begin{equation}
\label{eq:lemmaDisDir1}
\boldsymbol{\theta}_{r,w,(2:n_{\mathrm{clas}}),h}=\argmin_{\hat{\boldsymbol{\theta}}_{(2:n_{\mathrm{clas}}),h}}\Big(\sum_{c=2}^{n_{\mathrm{clas}}}\frac{1}{2}(\hat{\theta}_{c,h}-\tilde{\theta}_{c,h})^2+\frac{\lambda_{r,\mathrm{lda}}}{{[\boldsymbol{\Sigma}_{r,w}]}_{h,h}}{\|\hat{\boldsymbol{\theta}}_{(2:n_{\mathrm{clas}}),h}\|}_2\Big).
\end{equation}
The solution of \eqref{eq:lemmaDisDir1} was given by
\begin{equation}
\label{eq:lemmaDisDir2}
\boldsymbol{\theta}_{r,w,(2:n_{\mathrm{clas}}),h}=\tilde{\boldsymbol{\theta}}_{(2:n_{\mathrm{clas}}),h}\odot\mathrm{ReLU}\Big(1-\frac{\lambda_{r,\mathrm{lda}}}{{\|\tilde{\boldsymbol{\theta}}_{(2:n_{\mathrm{clas}}),h}\|}_2}\Big).
\end{equation}
\end{minipage}}
\end{center}
\end{figure}

As mentioned earlier, the inclusion of the Lasso regularizer in the objective function of the multivariate linear discriminant resulted in a nonlinearly penalized linear least squares estimator of no closed form solution. In \cite{Pan2021}, an efficient iterative algorithm got proposed to find the solution of this nonlinear optimization. Interestingly, the computational complexity of this algorithm was similar to the complexity of computing the closed form solution of the linear optimization resulted from the ridge regularizer. Thus, it motivated a wide use of the Lasso regularizer and its variants including the group Lasso regularizer.

The feasibility of the group Lasso regularizer allowed the MSDA to extend the discriminant directions of a \textbf{binary} linear discriminant, introduced in \eqref{eq:MLD2}, to their \textbf{multiclass} counterparts. Then, the MSDA expressed these discriminant directions as the minimizers of a quadratic objective function derived from a convex formulation of the Bayes rule and penalized with the group Lasso regularizer \cite{Pan2021}. The definition of these discriminant directions and the corresponding optimization problem were
\begin{equation}
\label{eq:disDirMultSepCov}
\boldsymbol{\theta}_{r,w,c}=\boldsymbol{\beta}_{r,w,c}-\boldsymbol{\beta}_{r,w,1}=\big[\boldsymbol{\Sigma}_{r,w}^{-1}\cdot(\boldsymbol{\mu}_{r,w,c}-\boldsymbol{\mu}_{r,w,1})\big]\in\mathbb{R}^{n_{\mathrm{feat}}}
\end{equation}
\begin{equation}
\label{eq:disDirMSDA}
\begin{split}
{[\boldsymbol{\theta}_{r,w,c}]}_{c\in\{2,\cdots,n_{\mathrm{clas}}\}}=\argmin_{[\hat{\boldsymbol{\theta}}_{2},\cdots,\hat{\boldsymbol{\theta}}_{n_{\mathrm{clas}}}]}\Bigg(&\sum_{c=2}^{n_{\mathrm{clas}}}\Big[\frac{1}{2}\hat{\boldsymbol{\theta}}^T_{c}\cdot\boldsymbol{\Sigma}_{r,w}\cdot\hat{\boldsymbol{\theta}}_{c}-(\boldsymbol{\mu}_{r,w,c}-\boldsymbol{\mu}_{r,w,1})^T\cdot\hat{\boldsymbol{\theta}}_{c}\Big]\\
&+\lambda_{r,\mathrm{lda}}\cdot\sum_{h=1}^{n_{\mathrm{feat}}}{\|\hat{\boldsymbol{\theta}}_{(2:n_{\mathrm{clas}}),h}\|}_2\Bigg)
\end{split}
\end{equation}
with $\boldsymbol{\mu}_{r,w,c}\in\mathbb{R}^{n_{\mathrm{feat}}}$ and $\boldsymbol{\Sigma}_{r,w}\in\mathbb{R}^{n_{\mathrm{feat}}\times n_{\mathrm{feat}}}$ being defined in \eqref{eq:MLD_Params1} and \eqref{eq:MLD_Params3}, respectively.

The original MSDA solved \eqref{eq:disDirMSDA} by using Lemma 1. This algorithm had a computational complexity of $\mathcal{O}\big(n_{\mathrm{feat}}^2\big)$ and was based on a blockwise coordinate descent optimization. It was memory intensive due to the storage and retrieval of the entire of matrix $\boldsymbol{\Sigma}_{r,w}\in\mathbb{R}^{n_{\mathrm{feat}}\times n_{\mathrm{feat}}}$ for its computations \cite{Mai2019}. In case of having a high-dimensional feature space, this storage become expensive. However, in such a space, it was likely that many features were irrelevant or correlated. The group Lasso regularizer of the MSDA returned zero discriminant directions for these features. Thus, it was unnecessary to load the elements of $\boldsymbol{\Sigma}_{r,w}$ which referred to useless features. This formed a modified MSDA listed in Algorithm \ref{AlgoMSDA}. The storage demand of this algorithm got drastically reduced by avoiding to retrieve the entire of $\boldsymbol{\Sigma}_{r,w}$. Instead, the blockwise operation of the coordinate descent optimization was used to only retrieve one column of $\boldsymbol{\Sigma}_{r,w}$ in each optimization trial. Also, the feature selection capability of the group Lasso regularizer was used to identify elements of the retrieved column which contributed to nonzero discriminant directions. Then, only these elements were involved in the computations. These actions not only reduced the memory footprint of the algorithm but also reduced its computational complexity to $\mathcal{O}\big(n_{\mathrm{feat}}\big)$ \cite{Pan2021}.

\begin{algorithm}[!t]
\caption{Modified multiclass sparse discriminant analysis (MSDA)}
\KwIn{The regularization hyperparameter $\lambda_{r,\mathrm{lda}}\in\mathbb{R}_{+}$, the feature vector $\mathbf{f}_{r,j}={[f_{r,j,h}]}_h\in\mathbb{R}^{n_{\mathrm{feat}}}$ and the reference label $l_{r,j}\in\mathbb{L}$ of each sample $v_{r,j}\in\mathbb{D}_{r,w}$, the covariance matrix $\boldsymbol{\Sigma}_{r,w}\in\mathbb{R}^{n_{\mathrm{feat}}\times n_{\mathrm{feat}}}$ and the mean vector $\boldsymbol{\mu}_{r,w,c}={[\mu_{c,h}]}_h\in\mathbb{R}^{n_{\mathrm{feat}}}$.}
\vspace{3mm}
\KwOut{The discriminant directions ${[\boldsymbol{\theta}_{r,w,c}\in\mathbb{R}^{n_{\mathrm{feat}}}]}_{c\in\mathbb{L}}$.}
\vspace{3mm}
\textbf{Set}~$\boldsymbol{\delta}_c=\boldsymbol{\mu}_{r,w,c}-\boldsymbol{\mu}_{r,w,1}={[\delta_{c,h}]}_h\in\mathbb{R}^{n_{\mathrm{feat}}}$ for $c\in\mathbb{L}=\{1,\cdots,n_{\mathrm{clas}}\}$.\\
\vspace{2mm}
\textbf{Initialize}~${[\theta^{(0)}_{c,h}]}_{c\in\mathbb{L}}$ and $\tilde{\theta}^{(0)}_{c,h}=\frac{(\delta_{c,h}-\sum_{h'\neq h}\theta^{(0)}_{c,h}\cdot{[\boldsymbol{\Sigma}_{r,w}]}_{h',h})}{{[\boldsymbol{\Sigma}_{r,w}]}_{h,h}}$ to form $\tilde{\boldsymbol{\theta}}^{(0)}_{(2:n_{\mathrm{clas}}),h}={[\tilde{\theta}^{(0)}_{c,h}]}_{c\in\mathbb{L}}$.\\
\BlankLine
\vspace{1mm}
\For{iteration~$i=1,\cdots$}{
\For{$h\in\{1,\cdots,n_{\mathrm{feat}}\}$}{
\begin{equation}
\label{eq:computeAlgoMSDA}
{[\theta^{(i)}_{c,h}]}_{c\in\mathbb{L}}=\tilde{\boldsymbol{\theta}}^{(i-1)}_{(2:n_{\mathrm{clas}}),h}\odot\mathrm{ReLU}\Big(1-\frac{\lambda_{r,\mathrm{lda}}}{{\big\|\tilde{\boldsymbol{\theta}}^{(i-1)}_{(2:n_{\mathrm{clas}}),h}\big\|}_2}\Big).
\end{equation}
\begin{subequations}
\label{eq:updateAlgoMSDA}
\begin{align*}
&\tilde{\theta}^{(i)}_{c,h}=\frac{N\cdot\theta^{(i)}_{c,h}-\sum_{h'\neq h}\theta^{(i)}_{c,h'}\cdot\Big(\sum_{c=1}^{n_{\mathrm{clas}}}\big[\sum_{v_{r,j}\in\mathbb{D}_{r,w,c}}(f_{r,j,h'}-\mu_{c,h'})\cdot(f_{r,j,h}-\mu_{c,h})\big]\Big)}{\sum_{c=1}^{n_{\mathrm{clas}}}\big[\sum_{v_{r,j}\in\mathbb{D}_{r,w,c}}(f_{r,j,h}-\mu_{c,h})^2\big]}\\
&~~~~~N=(\sum_{c'\in\mathbb{L}}n_{r,w,c'})-n_{\mathrm{clas}}~~~~~~~~~\mathbb{D}_{r,w,c}=\{v_{r,j}\in\mathbb{D}_{r,w}|l_{r,j}=c\in\mathbb{L}\}\\
&~~~~~~~~~~~~~~~~~~~~~~~~~~\tilde{\boldsymbol{\theta}}^{(i)}_{(2:n_{\mathrm{clas}}),h}={[\tilde{\theta}^{(i)}_{c,h}]}_{c\in\mathbb{L}}.\tag{\ref{eq:updateAlgoMSDA}}
\end{align*}
\end{subequations}}}
At convergence return ${\Big[\boldsymbol{\theta}_{r,w,c}={[\theta^{(i)}_{c,h}]}_{h\in\{1,\cdots,n_{\mathrm{feat}}\}}\Big]}_{c\in\mathbb{L}}$ and $\boldsymbol{\theta}_{r,w,1}=\mathbf{0}$.
\label{AlgoMSDA}
\end{algorithm}

\section{Outline of Contributions}
\label{sec:OutContRandForest}
In this chapter, we proposed a \textbf{hierarchical quadratic random forest classifier} to enable a simultaneous multiclass classification of multiresolution multichannel samples. This discriminative classifier was generic and applicable to any classification task. However, we applied it to the multiclass segmentation (patchwise classification) of objects (tissues) on volumetric (2-channel) fat-water MR images. Also, with regards to the arguments provided in page~\pageref{bagging-boosting}, this ensemble classifier got built and trained according to a bagging scheme.

To train (optimize) and to evaluate the proposed forest, the entire set of the volumetric fat-water images got divided into the training, validation, and test subsets. From the training, validation, and test images, the multiresolution training, validation, and test patches got extracted to respectively form the multiresolution training, validation, and test samples. The multiresolution training samples were used to optimize the main parameters (projecting coefficients and thresholds) and the hyperparameters of the forest. The multiresolution validation samples were used to evaluate the optimization of these parameters. The multiresolution test samples were used to evaluate the overall performance of the optimized forest alone or followed by the graphs proposed in \cite{Fallah2017a,Fallah2018p,Fallah2019b,Fallah2021}.

The proposed random forest classifier expanded on the methods described in \autoref{sec:BackMotivRandForest} by
\begin{itemize}[leftmargin=*]
\item building a multiresolution pyramid of $n_{\mathrm{lay}}$ resolution layers to extract cubic multiresolution fat-water patches from the volumetric fat-water images.
\item applying resolution-specific overlaps to the patches of each resolution to obtain similar number of patches across different resolutions and to sample larger patches denser.
\item extracting each fat-water patch from a volumetric fat-water image and forming a sample by extracting local and contextual intra- and inter-channel features from this patch\footnote{A patch accompanied with its features was called a sample.}.
\item using \textbf{hierarchical decision trees} with each tree encoding each resolution of the pyramid into one of its resolution layers to enable a hierarchical (coarse-to-fine or parent-to-child) classification of the multiresolution samples according to the pyramid.
\item mitigating imbalanced class-sample distribution of the training samples and thus the classification biases towards their dominant classes by applying a multiclass synthetic minority oversampling technique (SMOTE) to the training samples of each resolution layer.
\item using a penalized multivariate linear discriminant based on the multiclass sparse discriminant analysis (MSDA) in each decision node of the forest with the objective function given by \eqref{eq:disDirMSDA} and the optimization process listed in Algorithm \ref{AlgoMSDA} to
\begin{itemize}[leftmargin=*]
\item enable a multiclass classification in spite of features correlations and class overlaps;
\item mitigate possible rank deficiency of the covariance matrix, curse of dimensionality (high-dimensional feature vectors), and overfitting by selecting most discriminant features through the group Lasso regularizer of the MSDA of each decision node.
\end{itemize}
\item applying the linear discriminant to squared features to achieve \textbf{quadratic decision boundaries} and thus a lower classification bias in the original (unsquared) feature space without facing with the higher complexity of a quadratic discriminant (see \autoref{fig:LinearQuadraticBoundary}).
\item using a resolution-specific regularization hyperparameter $\lambda_{r,\mathrm{lda}}\in\mathbb{R}_{+}$ for the MSDAs employed at each resolution layer $r\in\{1,\cdots,n_{\mathrm{lay}}\}$ of the forest to adapt this hyperparameter to the number and distribution of the samples received in that layer.
\item allowing to improve the classifications by additionally considering the spatial (neighborhood) and the hierarchical fine-to-coarse (child-to-parent) relationships of the patches over the multiresolution graphs proposed in \cite{Fallah2017a,Fallah2018p,Fallah2019b,Fallah2021}. To this end, we
\begin{itemize}[leftmargin=*]
\item processed the multiresolution validation/test patches by the trained (optimized) forest;
\item interpreted the classification probabilities estimated by the forest as priors;
\item derived classification reliabilities of the multiresolution validation/test patches from the label heterogeneities of the multiresolution training patches processed by the forest;
\item consolidated the features selected by the group Lasso regularizers of the MSDAs of the decision nodes of the forest into resolution-specific and resolution-independent features to be used by the resolution-specific and resolution-independent graphs, respectively;
\item accompanied each patch with its resolution-specific features, its resolution-independent features, its priors, its spatial relationships, its bidirectional (parent-to-child and child-to-parent) hierarchical relationships, and its classification reliability to form a sample;
\item processed the multiresolution validation/test samples by the multiresolution graphs to estimate their classification posteriors and labels.
\end{itemize}
\end{itemize}

The local and contextual intra- and inter-channel features were designed to enhance the discriminative power of a classifier addressing an image segmentation task by capturing various textural patterns of the image regions. They included size- and rotation-variant as well as size- and rotation-invariant features. This was because the size and orientation of an image region of constant textural pattern were important information for segmenting objects (tissues) on it. However, these characteristics could be easily confused by the overall size and orientation of the image itself. Thus, we left it to the classifier to find its useful (discriminant) features. For a fat-water patch $\rho_{r,j}$ forming the sample $v_{r,j}$ of resolution $r\in\{1,\cdots,n_{\mathrm{lay}}\}$, the vector of the local and contextual intra- and inter-channel features was denoted by $\mathbf{f}''_{r,j}\in\mathbb{R}^{n_{\mathrm{tot}}}$ with $\mathrm{dim}(\mathbf{f}''_{r,j})=n_{\mathrm{tot}}$.

As described in page~\pageref{indWeakClass}, our proposed forest was collected according to a bagging scheme. That is, its classification variance (overfitting or prediction error on unseen samples) got minimized by diversifying its weak classifiers. This was achieved by feeding a randomly selected subset of the training samples and features to each weak classifier and optimizing its parameters independently from other weak classifiers. To this end, for each weak classifier, it was suggested to randomly select $\sqrt{n_{\mathrm{tot}}}$ features without replacement from the overall $n_{\mathrm{tot}}$ features \cite{Breiman2001}. Thus, the number of the weak classifiers $n_{\mathrm{weak}}$ should fulfill
\begin{equation}
\sqrt{n_{\mathrm{tot}}}\cdot n_{\mathrm{weak}}\leq n_{\mathrm{tot}}\implies n_{\mathrm{weak}}\leq\sqrt{n_{\mathrm{tot}}}=n_{\mathrm{sel}}.
\label{eq:numWeakClassCond}
\end{equation}
In this regard, we used $n_{\mathrm{weak}}=n_{\mathrm{sel}}$ weak classifiers in our forest and formed each weak classifier by a hierarchical decision tree. Then, for all the decision nodes of each weak classifier/tree we used $n_{\mathrm{sel}}$ randomly selected features which were different from the features used by other weak classifiers/trees. For a fat-water patch $\rho_{r,j}$ forming the sample $v_{r,j}$ of resolution $r\in\{1,\cdots,n_{\mathrm{lay}}\}$, these $n_{\mathrm{sel}}$ features were denoted by $\mathbf{f}'_{r,j}=[f'_{r,j,1},\cdots,f'_{r,j,n_{\mathrm{sel}}}]$.

To achieve quadratic decision boundaries and thus a lower classification bias in the space of $\mathbf{f}'_{r,j}\in\mathbb{R}^{n_{\mathrm{sel}}}$, as depicted in \autoref{fig:LinearQuadraticBoundary}, we formed squared features $\mathbf{f}_{r,j}\in\mathbb{R}^{n_{\mathrm{feat}}}$ from $\mathbf{f}'_{r,j}\in\mathbb{R}^{n_{\mathrm{sel}}}$ and processed these squared features by the multivariate linear discriminant of the node receiving that sample. For a fat-water patch $\rho_{r,j}$ forming the sample $v_{r,j}$ of resolution $r\in\{1,\cdots,n_{\mathrm{lay}}\}$, the vector of squared features $\mathbf{f}_{r,j}={[f_{r,j,h}]}_h\in\mathbb{R}^{n_{\mathrm{feat}}}$ was
\begin{equation}
\label{eq:squaredFeats}
\begin{split}
\mathbf{f}_{r,j}=&\Big[f'_{r,j,1},f'_{r,j,2},\cdots,f'_{r,j,n_{\mathrm{sel}}},\\
&(f'_{r,j,1}\cdot f'_{r,j,2}),(f'_{r,j,1}\cdot f'_{r,j,3}),\cdots,(f'_{r,j,22}\cdot f'_{r,j,n_{\mathrm{sel}}}),\\
&(f'^2_{r,j,1}),(f'^2_{r,j,2}),\cdots,(f'^2_{r,j,n_{\mathrm{sel}}})\Big]\\
n_{\mathrm{feat}}=\mathrm{dim}(\mathbf{f}_{r,j})=&~n_{\mathrm{sel}}+\binom{n_{\mathrm{sel}}}{2}+n_{\mathrm{sel}}=2\cdot n_{\mathrm{sel}}+\binom{n_{\mathrm{sel}}}{2}.
\end{split}
\end{equation}
That is, the vector of squared features not only involved the squares of the features but also the original (unsquared) features. During the training (optimization) of the forest, over the set of squared features fed to each decision node, the group Lasso regularizer of the MSDA of the node determined \textbf{indices of the most discriminant squared features}. This was called the \textbf{feature learning} capability of the forest. The selected squared features depended on the squared features and the compositions (voxelwise label histograms) of the training patches fed to the node in the training phase. Thus, their number and types (indices) could vary across the decision nodes of the forest or they were \textbf{node-specific}. Moreover, the selected features depended on the classification task and the quality, modality, and contrast of the training patches. For example, features selected for segmenting fatty tissues on fat-water MR images were not necessarily useful for detecting edema on T2-weighted MR images.
Furthermore, during the training (optimization) of the forest, each decision or leaf node inferred some empirical probabilities from the distribution of the training patches fed to it. These probabilities got assigned as the classification probabilities to each validation/test patch which visited the node in the validation/test phase. The probabilities were interpreted according to the desired classification accuracies. If the estimates of the forest were accurate enough then these probabilities were interpreted to be the \textbf{final classification probabilities} or the \textbf{posteriors}. If the estimates of the forest were supposed to be improved by another classifier such as the multiresolution graphs proposed in \cite{Fallah2017a,Fallah2018p,Fallah2019b,Fallah2021} then these probabilities were interpreted to be the \textbf{initial classification probabilities} or the \textbf{priors}.

To process the validation/test patches by a graph, all of these patches should be accompanied with a common set of feature types (indices). Thus, the \textbf{node-specific} features selected by the decision nodes of the forest could not be directly used by the graph. They should be first consolidated into a common set of feature types (indices) for all the patches.

The proposed multiresolution graphs involved a stack of neighborhood graphs ${\{\mathcal{G}_r\}}_{r=0}^{n_{\mathrm{lay}}}$ and the graph $\mathcal{G}_{\mathrm{hcrf}}$ of a hierarchical conditional random field (HCRF). Each neighborhood graph $\mathcal{G}_r$ processed patches of a certain resolution $r\in\{0,\cdots,n_{\mathrm{lay}}\}$ and the graph $\mathcal{G}_{\mathrm{hcrf}}$ processed patches of all the resolutions. Accordingly, for each neighborhood graph $\mathcal{G}_r$ we consolidated the node-specific features into a set of \textbf{resolution-specific} features and for the graph $\mathcal{G}_{\mathrm{hcrf}}$ we consolidated the node-specific features into a set of \textbf{resolution-independent} features.

In \cite{Mai2019,Pan2021}, the MSDA was proposed as a standalone classifier and, to the best of our knowledge, no previous method has used it as part of its ensemble. Thus, our proposed forest was the first ensemble classifier employing this discriminant in its decision nodes. Also, to the best of our knowledge, no previous work has processed squared features by a linear discriminant in order to achieve quadratic decision boundaries and thus a lower classification bias in the original (unsquared) feature space. In \cite{Gao2019}, a so-called quadratic forest was the result of multiple applications of the same (linear) forest to the input data.

As described in \autoref{sec:BackMotivRandForest}, rank deficiency of the covariance matrices and the curse of dimensionality could happen when the number of training samples were not enough or even less than the number of features. These effects become more probable by moving from the root node to the leaves of a nonhierarchical decision tree. This was because each decision node of this tree divided its received samples into $n_{\mathrm{clas}}$ subsets (classes) and then sent each subset to a decision node in the next (deeper) tree layer. Accordingly, by moving from the root node to the leaves of the tree, the number of the samples received by each decision node got reduced and the chance of having more features than the samples got increased.
The feature selection capability of the MSDA allowed to reduce the feature dimension and thus the above effects. We tackled these effects by not only using the MSDA in each decision node but also hierarchically decomposing the samples. To this end, each decision tree of the forest become hierarchical by encoding each resolution layer $r\in\{0,\cdots,n_{\mathrm{lay}}\}$ of the multiresolution pyramid proposed in \autoref{sec:PatchFeatExtr} into one of its resolution layers.

In the training phase, each hierarchical decision tree got recursively grown from its root node at its top to its leaves at its bottom. Thus, to be hierarchical, each tree received samples of the coarsest resolution $r=n_{\mathrm{lay}}$ from its root node and delivered classified voxels by its leaves at the finest resolution $r=0$. To this end, each decision node first classified its received samples and then decomposed each of these samples into its hierarchical children in the next (finer) resolution of the pyramid. Each child inherited the classification of its parent and according to this classification got sent to a decision node in the next (deeper) tree layer. This way, the number of samples received by each tree layer got increased by moving towards the deeper layers. This in turn reduced the chance of encountering with rank deficient covariance matrices, curse of dimensionality, or overfitting in the deeper tree layers due to having more features than samples there. In addition, it allowed a hierarchical (coarse-to-fine) classification of the multiresolution samples by considering their hierarchical relationships in a coarse-to-fine direction. In an image segmentation task, this enabled a fast localization of the addressed objects (tissues) in the coarser resolutions and an accurate segmentation of them in the finer resolutions. Moreover, it enhanced the reliability and robustness of the classifications in the presence of noise, artifacts, and outliers by enhancing the coarse-to-fine consistency of the classification probabilities estimated at different resolutions.

\begin{figure}[t!]
\begin{center}
\includegraphics[width=1.0\textwidth]{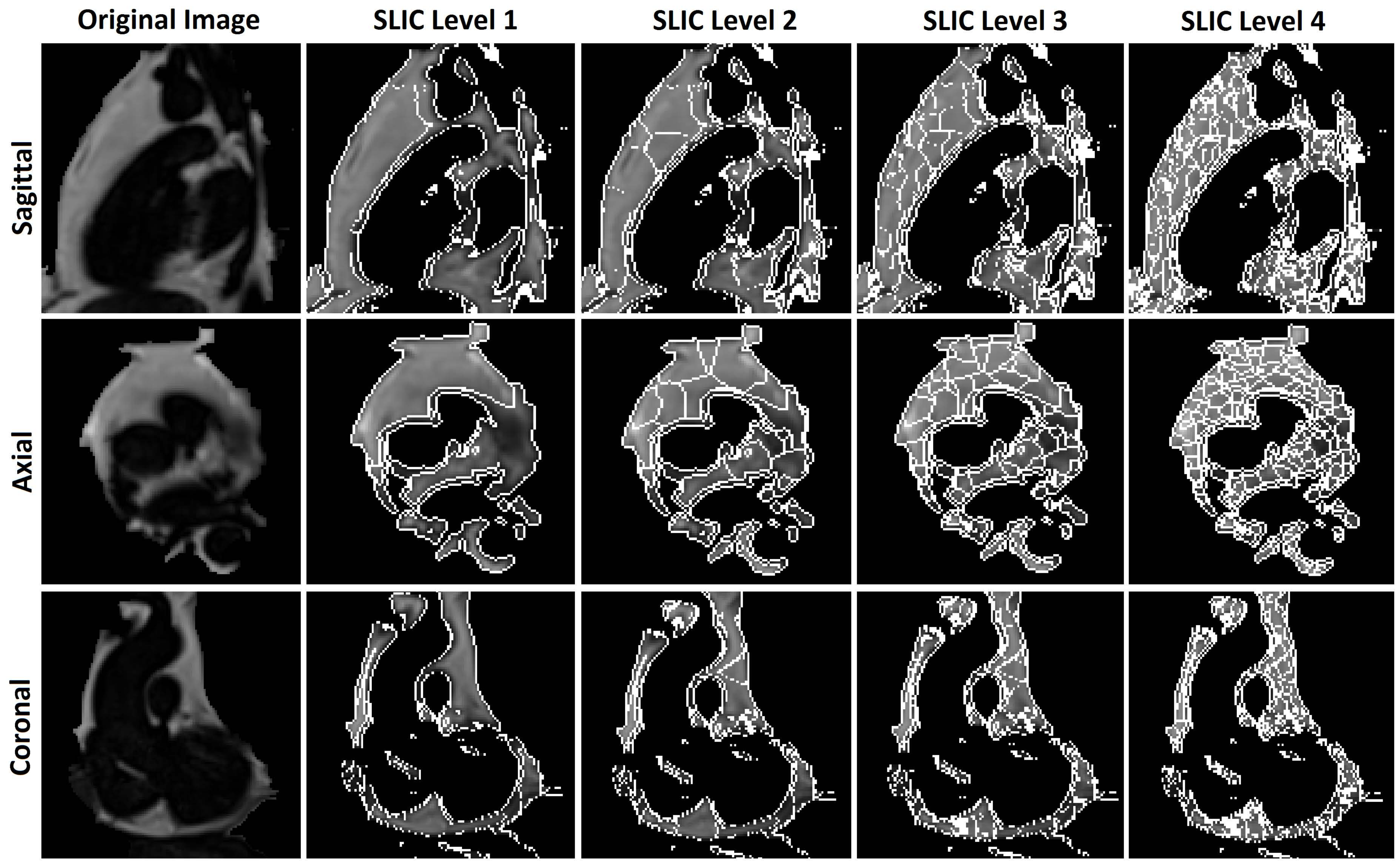}
\caption{SLIC clusters computed on a cardiac fat image by using different levels (cluster sizes).}
\label{fig:SLIC}
\end{center}
\end{figure}

In \cite{Geremia2013,Hutt2015,Zografos2016}, hierarchical classifiers got formed by encoding each resolution of a multiresolution pyramid into the entire of a decision tree. This implied a significantly higher memory footprint and computational complexity than our proposed forest which encoded each resolution of our pyramid into a resolution layer of each tree. In addition, those methods computed the samples' patches of each resolution by using an unsupervised simple linear iterative clustering (SLIC) technique of a certain level (resolution) \cite{Achanta2012}. As shown in \autoref{fig:SLIC}, the SLIC could be easily confused with the noise, artifacts, intensity nonuniformities, and similar textures of neighboring objects. These effects resulted in clusters which did not adhere to the objects boundaries.

The desired size (resolution) of the SLIC clusters was defined by a hyperparameter called \textit{level}. Despite of the definition of the \textit{level}, the clusters produced by the SLIC were irregularly shaped and sized. This made it hard to find an optimal \textit{level} for a segmentation task addressing objects of different shapes and/or sizes. To avoid these issues, our multiresolution pyramid was formed from cubic fat-water patches of regular shapes and sizes. Extraction of these patches needed no computation and also avoided the bottlenecks of an unsupervised clustering technique. Moreover, the cubic patches could be regularly and reversibly decomposed across different resolutions of the pyramid to form the multiresolution patches/samples. This resulted in decision trees of consistent hierarchies despite of variations in intensities, textures, and compositions of the patches at different resolutions. This approach also simplified the computation of the number of samples per class in each resolution for an application of a class-sample balancing technique such as the SMOTE \cite{Chawla2002,Wang2009}.

Furthermore, regular decomposition of the cubic patches across the resolutions of the pyramid untangled parent-child relationships of the corresponding samples for their hierarchical classifications over the resolution layers of the hierarchical trees.

\section{Formation of Multiresolution Patches}
\label{sec:PatchFeatExtr}
From every volumetric fat-water image, multiresolution patches got extracted according to a pyramid of $n_{\mathrm{lay}}$ resolution layers. In the $r^{\mathrm{th}}$ resolution layer of this pyramid, cubic patches of $\big(3\times2^{r-1}\big)^3$ voxels got extracted from every fat-water image. Accordingly, $r=n_{\mathrm{lay}}$ and $r=0$ represented the coarsest and the finest resolution, respectively. The $r=0$ was a special resolution layer formed by single-voxel patches and devised for the leaf nodes of the forest. \autoref{table:numberOfVoxelPerPatch} shows the number of voxels per patch at different resolutions of the pyramid.

\begin{table}[t!]
\begin{center}
\caption{Number of voxels per patch at different resolutions of the pyramid with $n_{\mathrm{lay}}=5$ layers.}
\label{table:numberOfVoxelPerPatch}
\resizebox{\textwidth}{!}{%
\begin{tabular}{|c|c|c|c|c|c|c|}
\hline
\textbf{Resolution Layer (r)}&5&4&3&2&1&0\\\hline
\#\textbf{Voxels per Patch}&$48^3=110592$&$24^3=13824$&$12^3=1728$&$6^3=216$&$3^3=27$&$1$\\\hline
\end{tabular}}
\end{center}
\end{table}

By considering the ratios of the images' sizes to the average size of the addressed objects (tissues), $n_{\mathrm{lay}}=5$ could enable a fast localization of the objects in the coarsest resolution $r=n_{\mathrm{lay}}$ without compromising their segmentation accuracies in the finest resolution $r=0$.

In each resolution layer $r\in\{1,\cdots,n_{\mathrm{lay}}\}$ of the pyramid, the fat-water patches of each image had $100\times\big(1-2^{(1-r)}\big)\%$ overlap with each other in each direction. This resolution-specific overlap was proportional to the patches' sizes and aimed to sample larger patches denser.

In an image, it was likely that patches of the same semantics be spatially close to each other. In addition, some fatty/lean tissues on an MR image were likely to be surrounded by certain lean/fatty tissues. This motivated to extract features of each patch not only from its spatial region but also from its neighboring patches. Features extracted from a patch region/neighborhood were referred to be local/contextual. A first-order neighborhood involved 4 direct spatial neighbors in 2D and 6 direct spatial neighbors in 3D. A second-order neighborhood involved 8 direct spatial neighbors in 2D and 26 direct spatial neighbors in 3D. The first-order neighborhood only covered the edges but the second-order neighborhood covered the edges and the corners between different patches. This allowed it to capture subtle variations across different patches \cite{Grady2008}. Accordingly, we used the 26-connected neighborhood in which each fat-water patch had 26 direct spatial neighbors.

Each fat-water patch had a fat and a water intensity channel. The intensity of the fat/water channel of each patch was the median\footnote{Median was selected due to its robustness to outliers caused by noise and artifacts.} of its voxelwise fat/water intensities. Based on these, from the fat or the water channel of each fat-water patch some basic features got extracted. These features included some kernel-based features whose extraction demanded the patch to be composed of subpatches called cells. Accordingly, we designed each patch of each resolution $r\in\{1,\cdots,n_{\mathrm{lay}}\}$ to be composed of $27$ cells in a $3\times 3\times 3$ stencil. This implied that the kernel-based features could not be extracted from the single-voxel patches of the resolution $r=0$. The only basic features of each patch in this resolution were its fat and water intensities.

\begin{figure}[t!]
\begin{center}
\includegraphics[width=1.0\textwidth]{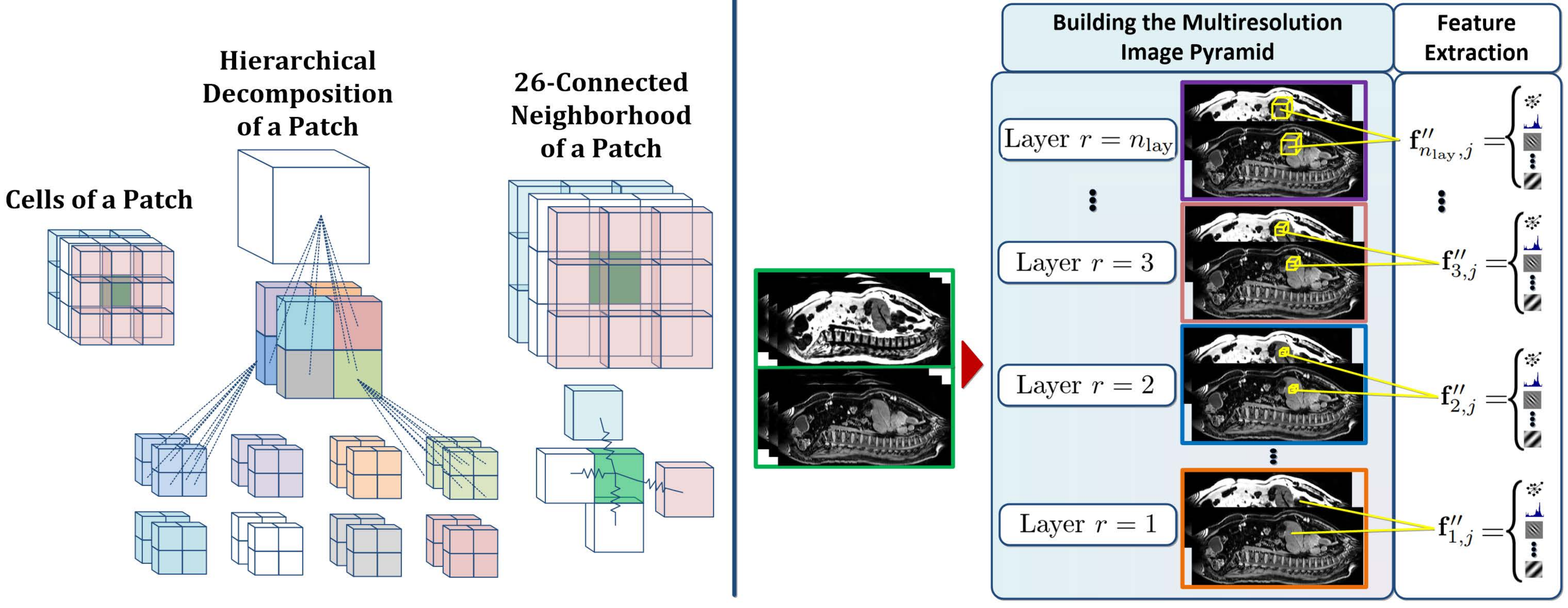}
\caption{a) Cells, hierarchical decomposition to $2^3=8$ children, and the 26-connected neighborhood of a patch. b) Feature extraction over the multiresolution pyramid.}
\label{fig:HierNeighFeatExtrct}
\end{center}
\end{figure}

From the basic features of each fat-water patch and its 26 neighbors, its local and contextual intra- and inter-channel features were computed. The local intra-channel features were from the fat \textbf{or} the water channel of the patch. The local inter-channel features were from the fat \textbf{and} the water channel of the patch. The contextual intra-channel features were from the fat \textbf{or} the water channel of the patch and its 26 neighbors. The contextual inter-channel features were from the fat/water channel of the patch \textbf{and} the water/fat channel of its 26 neighbors to measure the distribution of lean/fatty tissues around fatty/lean tissues.

The local and contextual intra- and inter-channel features of each fat-water patch $\rho_{r,j}$ formed a vector $\mathbf{f}''_{r,j}\in\mathbb{R}^{n_{\mathrm{tot}}}$. Then, to train the forest, as given by \eqref{eq:numWeakClassCond}, out of the $n_{\mathrm{tot}}$ features $\mathbf{f}''_{r,j}\in\mathbb{R}^{n_{\mathrm{tot}}}$, $n_{\mathrm{sel}}$ features $\mathbf{f}'_{r,j}\in\mathbb{R}^{n_{\mathrm{sel}}}$ got randomly selected for the weak classifier/tree processing the patch $\rho_{r,j}$.

Also, in summary, each fat-water patch $\rho_{r,j}$ of each resolution $r\in\{1,\cdots,n_{\mathrm{lay}}\}$ was composed of $27$ cells and had a hierarchical parent in the $(r+1)^{\mathrm{th}}$ resolution layer, $2^3=8$ hierarchical children in the $(r-1)^{\mathrm{th}}$ resolution layer, and $26$ direct spatial neighbors in its resolution layer $r$. \autoref{fig:HierNeighFeatExtrct} shows these and the feature extraction over the pyramid.

Furthermore, each fat-water patch $\rho_{r,j}$ had a \textbf{voxelwise label histogram} collected from the voxelwise reference labels obtained from the manual segmentations of the fat-water images. Mode of this histogram defined the \textbf{reference classification label} $l_{r,j}\in\mathbb{L}=\{1,\cdots,n_{\mathrm{clas}}\}$ of the patch. Its full width at half maximum (FWHM) $h_{r,j}$ reflected the \textbf{voxelwise label heterogeneity} of the patch\footnote{The FWHM of a Gaussian distribution was $2\cdot\sqrt{2\cdot\mathrm{ln}(2)}\cdot\sigma$ with $\sigma$ being the variance of the distribution. Thus, the FWHM was an easy-to-measure parameter in proportion to the variance.} which could stem from the existence of different objects/tissues in it or the effects of noise and artifacts on the manual segmentations. Thus, the voxelwise label heterogeneity $h_{r,j}$ of each patch $\rho_{r,j}$ indicated the \textbf{difficulty of its classification}.

The voxelwise label histogram of each training patch was included in its corresponding training sample. These samples were then used to optimize the parameters of the proposed forest according to the process described in \autoref{sec:TrainRF}. The voxelwise label histogram of each validation/test patch was not included in its corresponding validation/test sample. Only the reference classification labels of the validation/test patches were used to evaluate the proposed forest alone or followed by the graphs proposed in \cite{Fallah2017a,Fallah2018p,Fallah2019b,Fallah2021}.

\section{Formation of Multiresolution Training Samples}
\label{sec:MultResTrainSmpls}
The fat-water patch $\rho_{r,j}$ of a training fat-water image formed a training sample $v_{r,j}\in\mathbb{D}_{r,\mathrm{train}}$ of resolution $r\in\{1,\cdots,n_{\mathrm{lay}}\}$ by being accompanied with its
\begin{enumerate}[label={(\arabic*)},leftmargin=*]
\item\label{TrainStruct}hierarchical parent in the $(r+1)^{\mathrm{th}}$ resolution layer
\item $2^3=8$ hierarchical children in the $(r-1)^{\mathrm{th}}$ resolution layer
\item vector of squared features $\mathbf{f}_{r,j}\in\mathbb{R}^{n_{\mathrm{feat}}}$ formed according to \eqref{eq:squaredFeats} from the elements of $\mathbf{f}'_{r,j}\in\mathbb{R}^{n_{\mathrm{sel}}}$ selected out of the elements of $\mathbf{f}''_{r,j}\in\mathbb{R}^{n_{\mathrm{tot}}}$ for the tree processing it
\item reference label $l_{r,j}\in\mathbb{L}=\{1,\cdots,n_{\mathrm{clas}}\}$ (mode of its voxelwise label histogram)
\item voxelwise label heterogeneity $h_{r,j}$ (FWHM of its voxelwise label histogram) reflecting the difficulty of its classification.
\end{enumerate}

To make the features coherent and to stabilize the MSDA of the multivariate linear discriminant of each decision node, across all the training samples visiting the node, each element of the squared features got normalized to zero mean and unit variance \cite{Menze2011}.

\begin{figure}[t!]
\begin{center}
\includegraphics[width=0.9\textwidth]{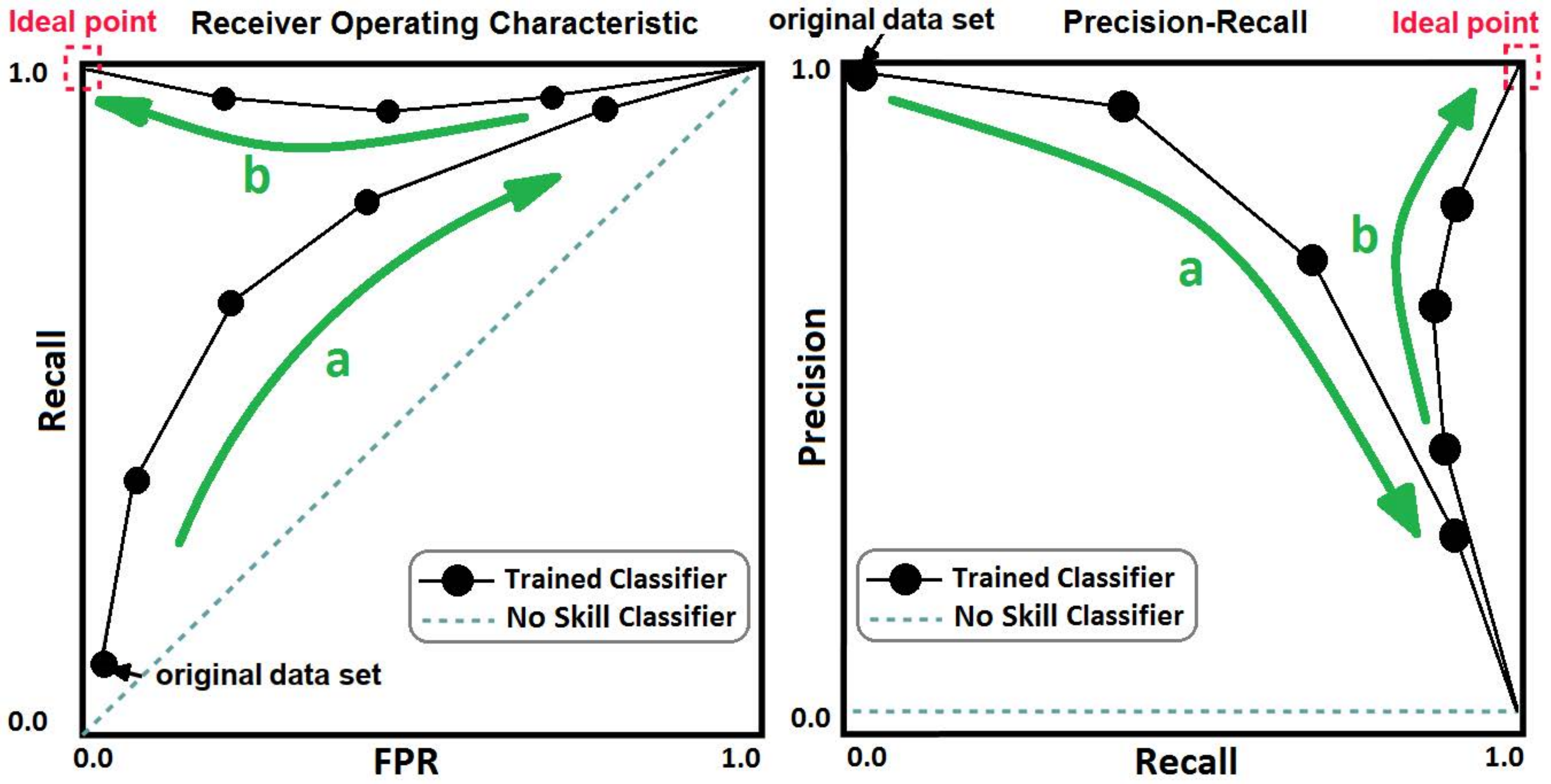}
\caption{Effects of the SMOTE on the performance of a classifier acting on a data set of imbalanced class-sample distribution. a) Transition of the performance by an increased undersampling of the samples of the majority (dominant) classes. b) Transition of the performance by an increased oversampling of the samples of the minority classes.}
\label{fig:ROC_PR_SMOTE}
\end{center}
\end{figure}

Most of the voxels in every fat-water image belonged to the background (unimportant) class(es). This led to an imbalanced class-sample distribution and produced classification biases towards dominant background classes. To mitigate these biases, we class-balanced the training samples of each resolution layer of each weak classifier/tree by applying a multiclass synthetic minority oversampling technique (SMOTE) to these samples. This technique downsampled the samples of the majority class by removing some of them. It also upsampled the samples of each minority class by generating some synthetic samples through an interpolation of the features of the realistic samples of the minority class \cite{Chawla2002,Wang2009}.

Effects of the undersampling and oversampling of the SMOTE on the performance of a classifier could be depicted by transitions over the \textit{receiver operating characteristic} and the \textit{precision-recall} curves. \autoref{fig:ROC_PR_SMOTE} shows these transitions.
The threshold metrics of \textit{recall}, \textit{specificity}, and \textit{precision} allowed to focus on a minority class to mitigate the classification biases towards the majority classes. Thus, in dealing with an imbalanced class-sample distribution, the \textit{precision-recall} curve was more reliable than the \textit{receiver operating characteristic} curve.

We produced the synthetic samples of each decision node of each tree independently from other nodes. That is, we did not propagate the synthetic samples across the resolution layers of each tree. This was because class-sample distribution of the samples received by each node could differ from other nodes. Also, the hierarchical decompositions of the realistic samples could produce children whose voxelwise label histograms and thus reference labels differed from their parents. This could lead to different class-sample distributions in different resolutions. Thus, the class-sample balancing should be done for each decision node in each resolution layer separately and independently from other layers and nodes.
In this regard, if the set of the realistic training samples received by a decision node $\eta_{r,w}$ in the resolution layer $r\in\{1,\cdots,n_{\mathrm{lay}}\}$ of the weak classifier/tree $w\in\{1,\cdots,n_{\mathrm{weak}}\}$ was $\mathbb{D}_{r,w}$, then the SMOTE was applied to this set and its set of feature vectors $\{\mathbf{f}'_{r,j}\in\mathbb{R}^{n_{\mathrm{sel}}}|v_{r,j}\in\mathbb{D}_{r,w}\}$. The subset of $\mathbb{D}_{r,w}$ containing the realistic samples with a reference classification label of $c\in\mathbb{L}$ was denoted by $\mathbb{D}_{r,w,c}=\{v_{r,j}\in\mathbb{D}_{r,w}|l_{r,j}=c\in\mathbb{L}\}$ with $|\mathbb{D}_{r,w,c}|=n_{r,w,c}$.

\begin{figure}[t!]
\begin{center}
\includegraphics[width=1.0\textwidth]{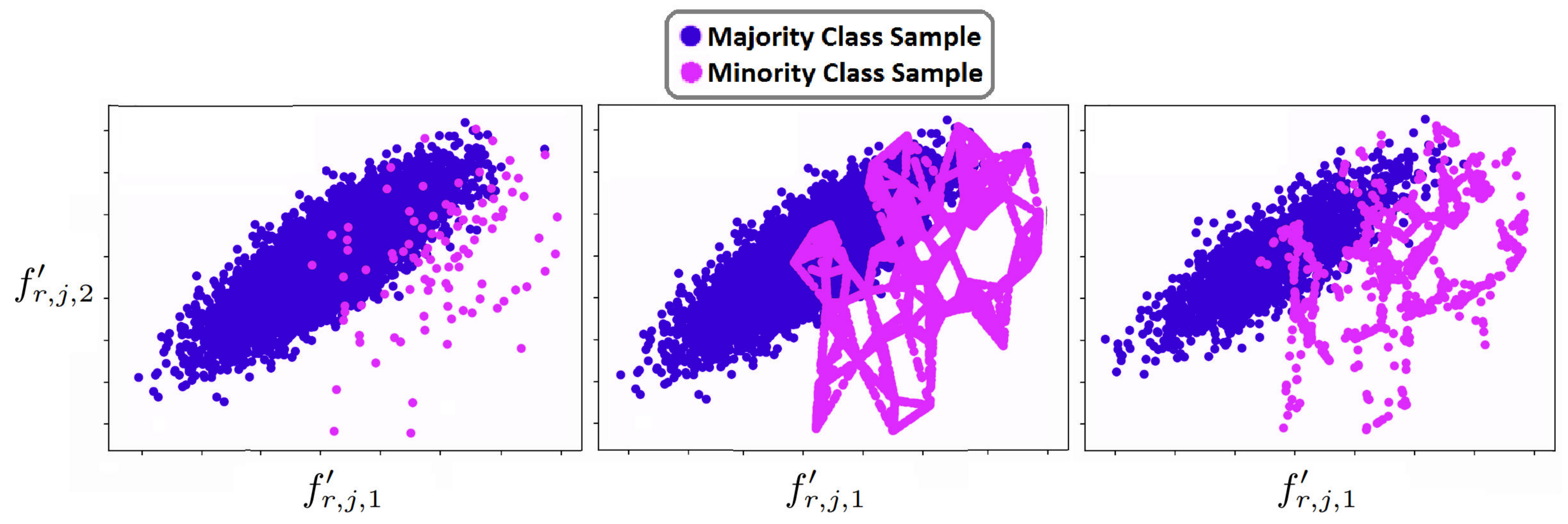}
\caption{\textbf{Left}) Samples of two classes after being projected into the plane of two most discriminant features ($f'_{r,j,1},~f'_{r,j,2}$). \textbf{Middle}) Results of interpolation (oversampling) of the samples of the minority classes by the SMOTE. \textbf{Right}) Results of removal (undersampling) of the samples of the majority classes by the SMOTE.}
\label{fig:SMOTE}
\end{center}
\end{figure}

We denoted the majority class of the highest number of realistic samples and the minority classes with $c'\in\mathbb{L}=\{1,\cdots,n_{\mathrm{clas}}\}$ and $\{\mathbb{L}-c'\}$, respectively. Thus, to achieve an almost balanced number of training samples per class, an oversampling rate of $100\times(n_{r,w,c'}/n_{r,w,c})\%$ was needed for every class label $c\in\{\mathbb{L}-c'\}$. To achieve this, the SMOTE randomly selected the samples of the majority class $c'\in\mathbb{L}$ without replacement and randomly selected the samples of every minority class $c\in\{\mathbb{L}-c'\}$ with $100\times\big(1-(n_{r,w,c}/n_{r,w,c'})\big)\%$ replacement. Then, it oversampled the selected realistic samples of every minority class by interpolating their feature vectors. Each interpolation was done in a neighborhood of the realistic samples in the feature space. More specifically, over a line connecting each realistic sample to one of its $n_{\mathrm{neigh}}\in\mathbb{N}$ nearest realistic neighbors, a \textbf{weighted average} of the feature vectors at the sides of the line gave the feature vector of a synthetic sample on the line.
The interpolation weight of the sample and its neighbor were $m_{w,\mathrm{int}}=w/(n_{\mathrm{weak}}+1)\in(0,1)$ and $(1-m_{w,\mathrm{int}})\in(0,1)$, respectively, with $w\in\{1,\cdots,n_{\mathrm{weak}}\}$ indicating the weak classifier/tree processing the samples. That is, the interpolation weight $m_{w,\mathrm{int}}\in(0,1)$ was specific to the $w^{\mathrm{th}}$ weak classifier/tree. This allowed to further diversify the training samples of different weak classifiers/trees and thus to enhance their dissimilarity and independence. Accordingly, each synthetic sample $s$ of each resolution $r\in\{1,\cdots,n_{\mathrm{lay}}\}$ had a feature vector $\mathbf{f}'_{r,s}\in\mathbb{R}^{n_{\mathrm{sel}}}$ from which its squared features $\mathbf{f}_{r,s}\in\mathbb{R}^{n_{\mathrm{feat}}}$ got computed. The squared features of the synthetic and realistic samples got used to optimize the parameters of the multivariate linear discriminant of each decision node receiving them in the training phase.

The number of nearest realistic neighbors $n_{\mathrm{neigh}}\in\mathbb{N}$ was defined by the oversampling rate. The oversampling rate for each class was the ratio of the number of realistic samples of the majority class to the number of realistic samples of this class. For example, if an oversampling rate of $200\%$ was needed for a minority class, then $n_{\mathrm{neigh}}=2$ nearest realistic neighbors of every realistic sample of this class should be used to generate two synthetic samples. \autoref{fig:SMOTE} shows the effects of the SMOTE on the samples of two imbalanced classes.

\section{Training of the Random Forest Classifier}
\label{sec:TrainRF}
\begin{figure}[t!]
\begin{center}
\includegraphics[width=1.0\textwidth]{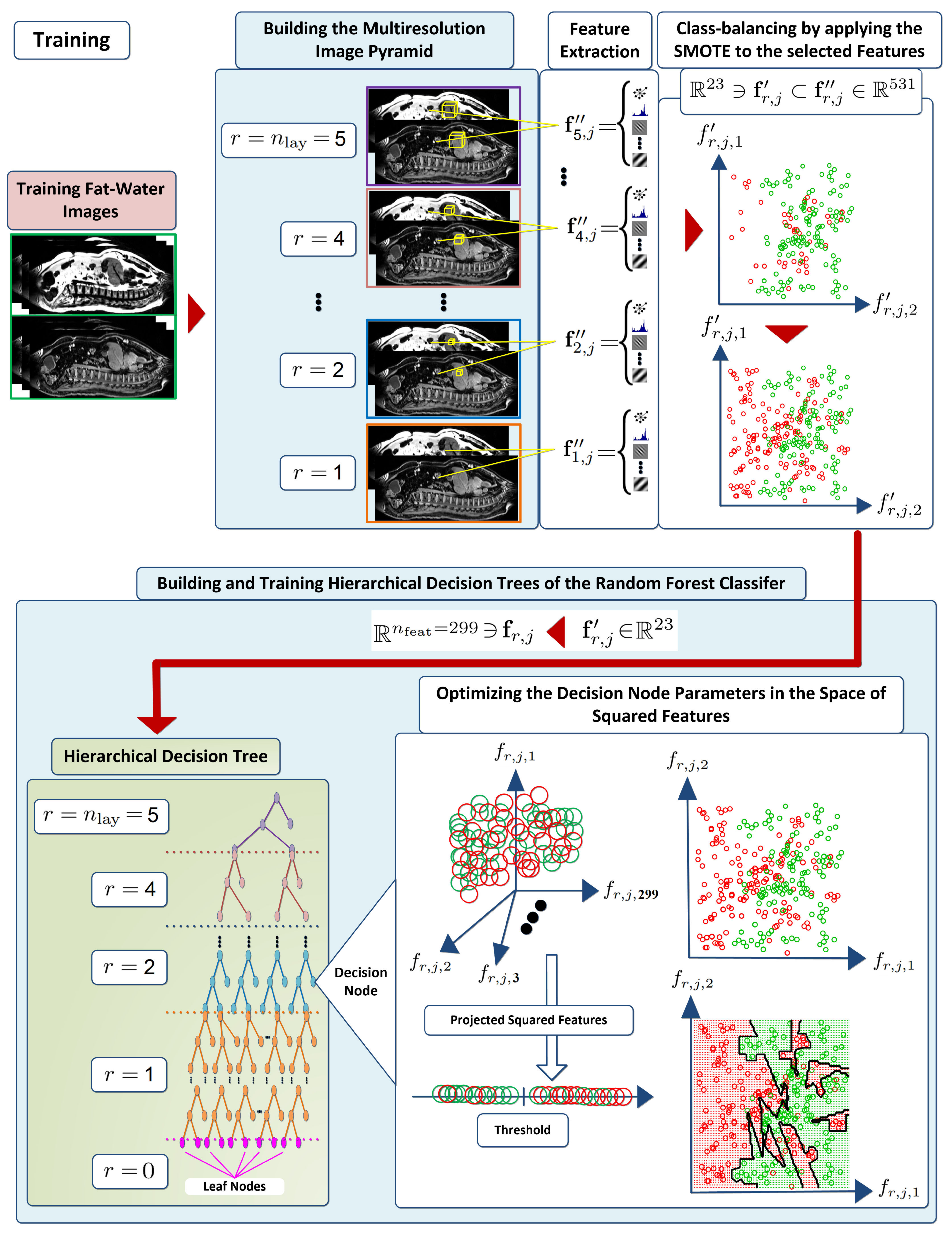}
\caption{The training process of the proposed random forest classifier.}
\label{fig:RandomForestTraining}
\end{center}
\end{figure}

As given by \eqref{eq:numWeakClassCond}, the proposed forest was composed of $n_{\mathrm{weak}}=n_{\mathrm{sel}}$ weak classifiers and each weak classifier had a hierarchical decision tree. The weak classifiers (decision trees) of this forest got built according to a bagging scheme. That is, they become diversified by feeding different subsets of the training samples and features to each of them and building each of them in parallel to (independent from) others. Also, as described in \autoref{sec:MultResTrainSmpls}, the training samples of each decision node of the proposed forest got class-balanced by applying the SMOTE to them. \autoref{fig:RandomForestTraining} shows the training process of the proposed forest.

In the training phase, each hierarchical decision tree got recursively grown from its root node at its top to its leaves at its bottom. To be hierarchical, each tree received samples of the coarsest resolution $r=n_{\mathrm{lay}}$ from its root node and delivered classified voxels by its leaves at the finest resolution $r=0$. To this end, each decision node at a resolution layer $r\in\{2,\cdots,n_{\mathrm{lay}}\}$ first classified its received samples and then decomposed each of these samples into its hierarchical children in the next (finer) resolution layer of the multiresolution pyramid proposed in \autoref{sec:PatchFeatExtr}. Each child inherited the classification of its parent and according to this classification got sent to a decision node in the next (deeper) tree layer. This way, for $r\in\{2,\cdots,n_{\mathrm{lay}}\}$, each resolution layer of the multiresolution pyramid got encoded into a physical layer of each hierarchical decision tree. However, in the resolution layer $r=1$, each tree could have as many physical layers as its \textbf{growth criterion} needed. The growth criterion in this resolution layer was based on the heterogeneity (\textbf{Gini impurity} or \textbf{entropy}) of the reference labels of the training samples received by each node in this layer.

Each decision node of the resolution layer $r=1$ did not decompose its received samples into a finer resolution. It only classified these samples and according to their classifications sent them to some other nodes in the next (deeper) physical layer of the same resolution layer unless it become a leaf node. Each node at the bottom of the resolution layer $r=1$ turned into a \textbf{leaf node} and thus stopped a further growth of the tree from its position if the Gini impurity of its received training samples was less than $g_{\mathrm{tree}}\in\mathbb{R}$. This way, in the resolution layer $r=1$, each tree got grown until it reached an overall depth of $d_{\mathrm{tree}}\in\mathbb{N}$ (excluding the leaf nodes) or a Gini impurity of less than $g_{\mathrm{tree}}\in\mathbb{R}$ in each of its leaf nodes. The maximum Gini impurity $g_{\mathrm{tree}}\in\mathbb{R}$ and the maximum tree depth $d_{\mathrm{tree}}\in\mathbb{N}$ were hyperparameters.

Each decision node $\eta_{r,w}$ of each resolution layer $r\in\{1,\cdots,n_{\mathrm{lay}}\}$ optimized its parameters by classifying its received \textbf{class-balanced training samples} in the set $\mathbb{D}_{r,w}$. As described in \autoref{sec:OutContRandForest}, each decision node was equipped with a penalized multivariate linear discriminant based on the MSDA. The optimization problem of this algorithm was given by \eqref{eq:disDirMSDA} and its optimization process was listed in Algorithm \ref{AlgoMSDA}. The MSDA got applied to the set of squared features $\{\mathbf{f}_{r,j}\in\mathbb{R}^{n_{\mathrm{feat}}}|v_{r,j}\in\mathbb{D}_{r,w}\}$ of the samples in $\mathbb{D}_{r,w}$. After the convergence, it returned the optimum discriminant direction $\boldsymbol{\theta}_{r,w,c}\in\mathbb{R}^{n_{\mathrm{feat}}}$ of each class $c\in\mathbb{L}=\{1,\cdots,n_{\mathrm{clas}}\}$. Then, based on \eqref{eq:MLD} and \eqref{eq:MLD2}, from $\boldsymbol{\theta}_{r,w,c}\in\mathbb{R}^{n_{\mathrm{feat}}}$ and the optimum projecting coefficients $\boldsymbol{\beta}_{r,w,1}=\boldsymbol{\Sigma}_{r,w}^{-1}\cdot\boldsymbol{\mu}_{r,w,1}$ of the reference class $(c=1)$, the optimum projecting coefficients of each class $c\in\mathbb{L}$ was obtained as $\boldsymbol{\beta}_{r,w,c}=\boldsymbol{\theta}_{r,w,c}+\boldsymbol{\beta}_{r,w,1}$.

In each vector $\boldsymbol{\beta}_{r,w,c}\in\mathbb{R}^{n_{\mathrm{feat}}}$, the group Lasso regularizer of the MSDA had already vanished the coefficients of unimportant (useless) squared features. By consolidating the indices of nonzero coefficients of all the $n_{\mathrm{clas}}$ vectors ${\{\boldsymbol{\beta}_{r,w,c}\in\mathbb{R}^{n_{\mathrm{feat}}}\}}_{c=1}^{n_{\mathrm{clas}}}$, indices of the most discriminant squared features for the classifications of the node $\eta_{r,w}$ were found.

The multiplication of the projecting coefficients with each $\mathbf{f}_{r,j}\in\mathbb{R}^{n_{\mathrm{feat}}}$ resulted in
\begin{equation}
\label{eq:projSpace}
[\mathbf{f}_{r,j}^T\cdot\boldsymbol{\beta}_{r,w,1},\cdots,\mathbf{f}_{r,j}^T\cdot\boldsymbol{\beta}_{r,w,n_{\mathrm{clas}}}]\in\mathbb{R}^{n_{\mathrm{clas}}}.
\end{equation}
In this lower-dimensional space, the multivariate linear discriminant of the node $\eta_{r,w}$ sought decision thresholds $\hat{\boldsymbol{\tau}}_{r,w}=[\hat{\tau}_{r,w,2},\cdots,\hat{\tau}_{r,w,n_{\mathrm{clas}}}]\in\mathbb{R}^{n_{\mathrm{clas}}-1}$ which could divide the samples in $\mathbb{D}_{r,w}$ into $n_{\mathrm{clas}}$ classes (subsets) denoted by $\{\hat{\mathbb{D}}_{r,w,c}\}_{c=1}^{n_{\mathrm{clas}}}$. These subsets fulfilled
\begin{equation}
\label{eq:divSetTrain}
\begin{split}
\hat{\mathbb{D}}_{r,w,1}\cup&~\hat{\mathbb{D}}_{r,w,2}\cup\cdots\cup\hat{\mathbb{D}}_{r,w,n_{\mathrm{clas}}}=\mathbb{D}_{r,w}\\
\hat{\mathbb{D}}_{r,w,1}\cap&~\hat{\mathbb{D}}_{r,w,2}=\hat{\mathbb{D}}_{r,w,2}\cap\hat{\mathbb{D}}_{r,w,3}=\cdots=\hat{\mathbb{D}}_{r,w,(n_{\mathrm{clas}}-1)}\cap\hat{\mathbb{D}}_{r,w,n_{\mathrm{clas}}}=\emptyset.
\end{split}
\end{equation}

The Gini impurity $g_{r,w}$ and the entropy $h_{r,w}$ of the set $\mathbb{D}_{r,w}$ were given by
\begin{equation}
\label{eq:giniImp}
g_{r,w}=g(\mathbb{D}_{r,w})=1-\sum_{c\in\mathbb{L}}\Bigg(\underbrace{\frac{\mathrm{card}\big(\{v_{r,j}\in\mathbb{D}_{r,w}|l_{r,j}=c\}\big)}{\mathrm{card}(\mathbb{D}_{r,w})}}_{\text{empirical probability}}\Bigg)^2=1-\sum_{c\in\mathbb{L}}(\underbrace{p_{r,w,c}}_{\substack{\text{empirical}\\\text{probability}}})^2
\end{equation}
\begin{equation}
\label{eq:empEntr}
h_{r,w}=h(\mathbb{D}_{r,w})=-\sum_{c\in\mathbb{L}}p_{r,w,c}\cdot\mathrm{log}_2(p_{r,w,c}).
\end{equation}
Due to the $\mathrm{log}_2$ operation, the computation of the entropy was more expensive than the Gini impurity. Thus, we only used the Gini impurity metric. The Gini impurity $g_{r,w}=g(\mathbb{D}_{r,w})$ got maximized when $\forall c\in\mathbb{L}:~p_{r,w,c}=1/n_{\mathrm{clas}}$. It become minimized $(0)$ when all the samples in $\mathbb{D}_{r,w}$ had the same reference label. Accordingly, $g_{r,w}\in[0,(n_{\mathrm{clas}}-1)/n_{\mathrm{clas}}]$.

The Gini impurity $\hat{g}_{r,w}$ of the divided set $\mathbb{D}_{r,w}=\hat{\mathbb{D}}_{r,w,1}\cup\cdots\cup\hat{\mathbb{D}}_{r,w,n_{\mathrm{clas}}}$ was
\begin{equation}
\label{eq:giniDivided}
\hat{g}_{r,w}=\hat{g}(\mathbb{D}_{r,w})=\sum_{c\in\mathbb{L}}\Bigg(\frac{\mathrm{card}(\hat{\mathbb{D}}_{r,w,c})}{\mathrm{card}(\mathbb{D}_{r,w})}\cdot g(\hat{\mathbb{D}}_{r,w,c})\Bigg)
\end{equation}
with the Gini impurity $g(\hat{\mathbb{D}}_{r,w,c})$ of each subset $\hat{\mathbb{D}}_{r,w,c}$ being
\begin{equation}
\label{eq:giniSubset}
g(\hat{\mathbb{D}}_{r,w,c})=1-\sum_{c'\in\mathbb{L}}\Bigg(\frac{\mathrm{card}\big(\{v_{r,j}\in\hat{\mathbb{D}}_{r,w,c}|l_{r,j}=c'\}\big)}{\mathrm{card}(\hat{\mathbb{D}}_{r,w,c})}\Bigg)^2.
\end{equation}
That is, the Gini impurity $\hat{g}_{r,w}$ of the divided set $\mathbb{D}_{r,w}$ was a weighted average of the Gini impurities of its subsets. The smaller the $\hat{g}_{r,w}$ was, the more close the estimated classification labels of the samples in $\mathbb{D}_{r,w}$ to their corresponding reference labels would be. Thus, $\hat{g}_{r,w}$ reflected the inaccuracy of the classifications of the node $\eta_{r,w}$. In this regard, the \textbf{optimum thresholds} $\boldsymbol{\tau}_{r,w}=[\tau_{r,w,2},\cdots,\tau_{r,w,n_{\mathrm{clas}}}]\in\mathbb{R}^{n_{\mathrm{clas}}-1}$ were
\begin{equation}
\label{eq:optThrshlds}
\boldsymbol{\tau}_{r,w}=\argmin_{\hat{\boldsymbol{\tau}}_{r,w}}~\hat{g}_{r,w}.
\end{equation}
The optimum thresholds were found by an exhaustive search over a set of discretized values. At the end of this optimization, the \textbf{minimized Gini impurity} of the divided set $\mathbb{D}_{r,w}$ was $\hat{g}^{*}_{r,w}$ and reflected the \textbf{inaccuracy} of the classifications of the decision node $\eta_{r,w}$.

As described in \autoref{sec:PatchFeatExtr}, in the resolution layer $r=1$ of the multiresolution pyramid, each fat-water patch was composed of $27$ voxels and had $100\times\big(1-2^{(1-1)}\big)\%=0\%$ overlap with other patches of the same resolution. In this resolution layer of each tree, each decision node $\eta_{1,w}$ turned into a leaf node $\eta_{0,w}$ if the Gini impurity $g_{1,w}=g(\mathbb{D}_{1,w})$ of its received training samples in $\mathbb{D}_{1,w}$ was less than the maximum Gini impurity $g_{\mathrm{tree}}\in\mathbb{R}$.

Every leaf node belonged to the special resolution layer $r=0$ in which each patch had a single voxel. In contrast to the decision nodes, each leaf node was not equipped with a discriminant. Thus, it could neither classify its received samples nor select/extract their features. Instead, each leaf node $\eta_{0,w}$ decomposed the patch of each of its received samples from $\mathbb{D}_{1,w}$ into $27$ voxels. All the resulting voxels formed the set $\mathbb{D}_{0,w}$ and inherited the classification label, namely $c'\in\mathbb{L}$, estimated by a decision node in the resolution layer $r=1$ for their parent samples in $\mathbb{D}_{1,w}$. Then, the leaf node computed the empirical probabilities $\mathbf{p}_{0,w}={[p_{0,w,c}]}_{c\in\mathbb{L}}$ and the Gini impurity $g_{0,w}=g(\mathbb{D}_{0,w})$ of the voxels in $\mathbb{D}_{0,w}$ by using their voxelwise reference labels. Since all the voxels in $\mathbb{D}_{0,w}$ had the same estimated label $c'\in\mathbb{L}$, the Gini impurity $\hat{g}^{*}_{0,w}=\hat{g}^{*}(\mathbb{D}_{0,w})$ of the divided set $\mathbb{D}_{0,w}=\hat{\mathbb{D}}_{0,w,c'}$ was
\begin{equation}
\label{eq:giniDividedLeaf}
\hat{g}^{*}_{0,w}=\hat{g}^{*}(\mathbb{D}_{0,w})=\frac{\mathrm{card}(\overbrace{\hat{\mathbb{D}}_{0,w,c'}}^{\mathbb{D}_{0,w}})}{\mathrm{card}(\mathbb{D}_{0,w})}\cdot g(\underbrace{\hat{\mathbb{D}}_{0,w,c'}}_{\mathbb{D}_{0,w}})=g(\mathbb{D}_{0,w})
\end{equation}
with the voxelwise Gini impurity $g_{0,w}=g(\mathbb{D}_{0,w})$ being
\begin{equation}
\label{eq:voxelwiseGiniLeaf}
g_{0,w}=g(\mathbb{D}_{0,w})=1-\sum_{c\in\mathbb{L}}\Bigg(\underbrace{\frac{\mathrm{card}\big(\{v_{r,j}\in\mathbb{D}_{0,w}|l_{r,j}=c\}\big)}{\mathrm{card}(\mathbb{D}_{0,w})}}_{\text{voxelwise empirical probability}}\Bigg)^2=1-\sum_{c\in\mathbb{L}}(\underbrace{p_{0,w,c}}_{\substack{\text{voxelwise}\\\text{empirical}\\\text{probability}}})^2.
\end{equation}

Similar to the decision nodes, the Gini impurity $\hat{g}^{*}_{0,w}=\hat{g}^{*}(\mathbb{D}_{0,w})$ reflected the \textbf{inaccuracy} of the voxelwise classifications of the leaf node $\eta_{0,w}$. However, it simply boiled down to the voxelwise Gini impurity $g_{0,w}=g(\mathbb{D}_{0,w})$ of this node. Consequently, the \textbf{classification accuracy} of each decision or leaf node $\eta_{r,w}$ of each resolution layer $r\in\{0,\cdots,n_{\mathrm{lay}}\}$ was reflected by $(1-\hat{g}^{*}_{r,w})$. The empirical probabilities computed by each decision/leaf node in the training phase got assigned to each patch which visited the node in the validation or test phase. These probabilities got interpreted according to the desired classification accuracies. If the classification estimates of the forest were accurate enough then these probabilities were interpreted to be the \textbf{final classification probabilities} or the \textbf{posteriors}. If the classification estimates of the forest were supposed to be improved by another classifier such as the multiresolution graphs proposed in \cite{Fallah2017a,Fallah2018p,Fallah2019b,Fallah2021} then these probabilities were interpreted to be the \textbf{initial classification probabilities} or the \textbf{priors}.

Furthermore, as described in \autoref{sec:MultResTrainSmpls}, each training sample $v_{r,j}$ of each resolution $r\in\{1,\cdots,n_{\mathrm{lay}}\}$ was accompanied with the voxelwise label heterogeneity $h_{r,j}$ of its fat-water patch $\rho_{r,j}$. This parameter was the FWHM of the voxelwise label histogram of the patch and reflected the variance of the patch composition and thus the difficulty of its classification.

During the training, each decision node $\eta_{r,w}$ measured the average of the voxelwise label heterogeneities of its received training samples in $\mathbb{D}_{r,w}$. This average was denoted by
\begin{equation}
\label{eq:avgVoxLabHet}
h_{r,w}=\frac{1}{|\mathbb{D}_{r,w}|}\cdot\sum_{v_{r,j}\in\mathbb{D}_{r,w}}h_{r,j}
\end{equation}
and reflected the \textbf{averaged difficulty of the classifications} of the decision node $\eta_{r,w}$. For each leaf node $\eta_{0,w}$, this average was $0$ because the label heterogeneity of each voxel was $0$.

\section{Parameters Saved in the Nodes of the Forest}
\label{sec:ParamsNodes}
At the end of the training, each decision node $\eta_{r,w}$ of the forest saved
\begin{enumerate}[label={(\arabic*)},leftmargin=*]
\item\label{ParsDecNode}indices of the squared features selected by the group Lasso regularizer of its MSDA;
\item the optimum projecting coefficients ${[\boldsymbol{\beta}_{r,w,c}\in\mathbb{R}^{n_{\mathrm{feat}}}]}_{c\in\mathbb{L}}$ and the optimum thresholds $\boldsymbol{\tau}_{r,w}\in\mathbb{R}^{n_{\mathrm{clas}}-1}$ of the MSDA of its multivariate linear discriminant;
\item the empirical probabilities $\mathbf{p}_{r,w}={[p_{r,w,c}]}_{c\in\mathbb{L}}$ of its received training samples;
\item its classification accuracy reflected by $(1-\hat{g}^{*}_{r,w})$;
\item the averaged difficulty of its classifications reflected by $h_{r,w}$.
\end{enumerate}

Similarly, each leaf node $\eta_{0,w}$ of the forest saved
\begin{enumerate}[label={(\arabic*)},leftmargin=*]
\item the empirical probabilities $\mathbf{p}_{0,w}={[p_{0,w,c}]}_{c\in\mathbb{L}}$ of its received training voxels;
\item its classification accuracy reflected by $(1-\hat{g}^{*}_{0,w})$;
\item the averaged difficulty of its classifications reflected by $h_{0,w}=0$\footnote{The $0$ value was not needed to be saved. It was declared for sake of consistency with the decision nodes.}.
\end{enumerate}

Each leaf node was preceded by at least one decision node of the resolution layer $r=1$. That is, if the Gini impurity of a decision node in the resolution layer $r=1$ was less than $g_{\mathrm{tree}}\in\mathbb{R}$ but this node was not preceded by any other decision node in the same resolution layer, then it did not turn into a leaf node and allowed the tree to grow further. As described in \autoref{sec:MultResValTestSmpls}, this was due to the fact that the information saved in the deepest decision/leaf node of each resolution layer $r\in\{0,\cdots,n_{\mathrm{lay}}\}$ of each decision tree were used to form the validation and test samples of this resolution. Thus, it was necessary to have at least one node in every resolution layer during each hierarchical traverse from the coarsest resolution to the finest resolution of each decision tree of the forest.

\section{Formation of Validation and Test Samples}
\label{sec:MultResValTestSmpls}
The proposed hierarchical random forest classifier could be used as a standalone classifier or followed by the multiresolution graphs proposed in \cite{Fallah2017a,Fallah2018p,Fallah2019b,Fallah2021} to improve its classifications. In both cases, the multiresolution training samples ${\{\mathbb{D}_{r,\mathrm{train}}\}}_{r=0}^{n_{\mathrm{lay}}}$ were used to optimize the main parameters (projecting coefficients and thresholds of the decision nodes) of the forest and the hyperparameters of the forest and the graphs proposed in \cite{Fallah2017a,Fallah2018p,Fallah2019b,Fallah2021}. The multiresolution validation samples ${\{\mathbb{D}_{r,\mathrm{val}}\}}_{r=0}^{n_{\mathrm{lay}}}$ were used to evaluate these optimizations. After the optimizations, the multiresolution test samples ${\{\mathbb{D}_{r,\mathrm{test}}\}}_{r=0}^{n_{\mathrm{lay}}}$ were used to evaluate the classification performance of the forest alone or followed by the graphs proposed in \cite{Fallah2017a,Fallah2018p,Fallah2019b,Fallah2021}.

The multiresolution validation/test samples got generated by extracting multiresolution fat-water patches from the volumetric fat-water images of the validation/test data sets and then extracting features from the patches. The patches' sizes (resolutions) and their hierarchical (parent-child) relationships with the patches of other resolutions were defined by the multiresolution pyramid proposed in \autoref{sec:PatchFeatExtr}. As given by \eqref{eq:numWeakClassCond}, the proposed forest had $n_{\mathrm{weak}}=n_{\mathrm{sel}}$ weak classifiers and each weak classifier had a hierarchical decision tree.

In contrast to the training patches, each validation/test patch was not only processed by a weak classifier/tree but by all the weak classifiers/trees in the forest. This allowed to collect as many classification estimates for it as possible. Also, the class-sample distribution of the validation/test patches did not get balanced by the SMOTE described in \autoref{sec:MultResTrainSmpls}. These were because the validation/test patches were not used in the optimizations (trainings). Thus, they played no role in the diversification of the weak classifiers/trees (by feeding different patches and features to them) during their optimizations nor in the classification biases towards the dominant classes of the training patches. Accordingly, the validation/test patches of the coarsest resolution got fed to the root node of every tree in the forest to be hierarchically classified and decomposed from the coarsest resolution to the finest resolution.

\begin{figure}[t!]
\begin{center}
\includegraphics[width=0.9\textwidth,height=20cm]{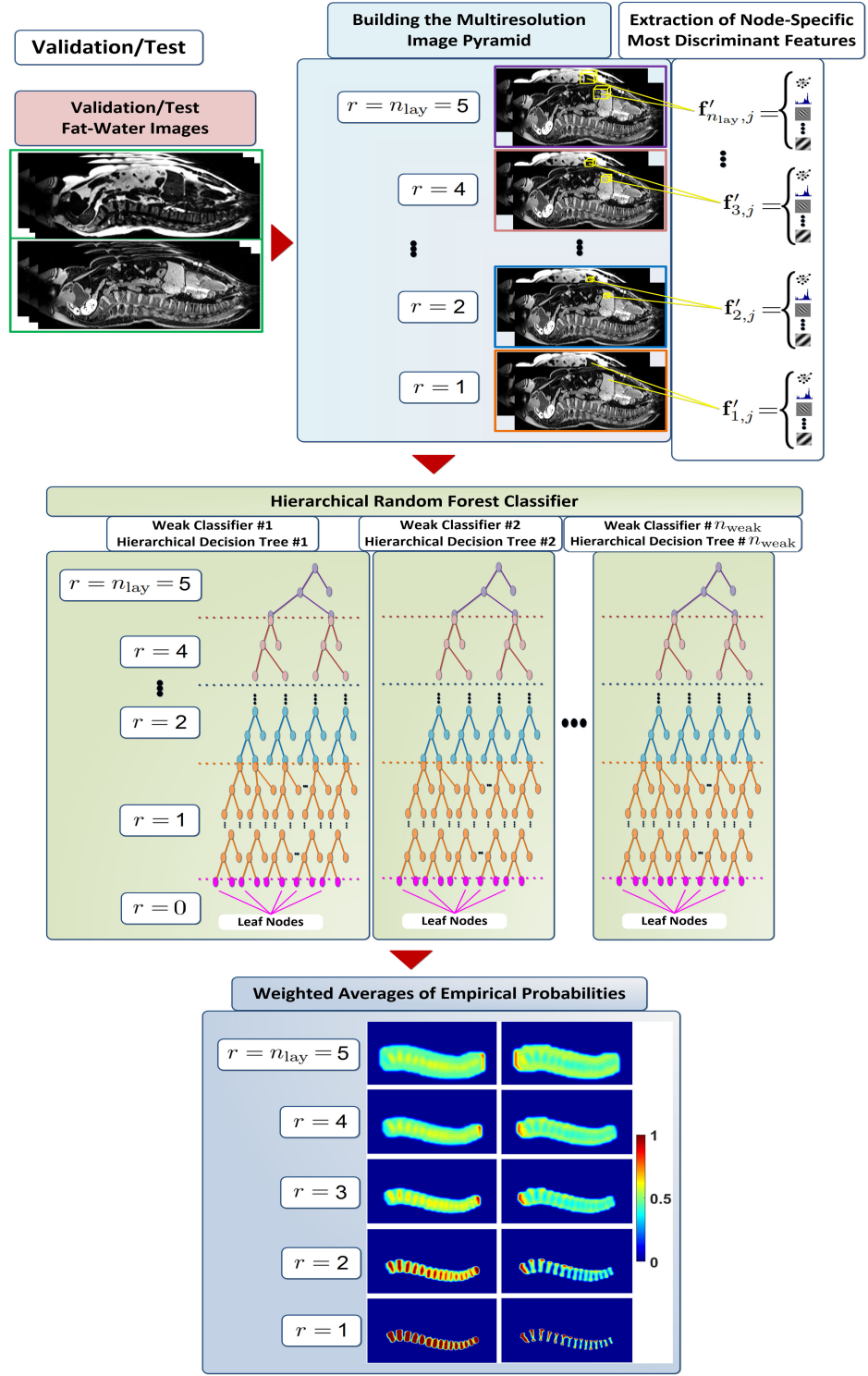}
\caption{Processing of a validation/test fat-water image by the proposed forest.}
\label{fig:RandomForestValidTest}
\end{center}
\end{figure}

However, to avoid computation of unnecessary features, in contrast to the training patches, features of each validation/test patch were not computed in advance rather on the fly when it visited a new decision node in the forest. That is, only the squared features whose indices were saved in the decision node and were indicated by (1) in the enumeration list of \autoref{sec:ParamsNodes} got extracted from the patch. To make the features coherent and to stabilize the linear discriminant of each decision node, similar to the training process, across all the validation/test patches visiting the node, each element of the extracted features got normalized to zero mean and unit variance. Then, the projecting coefficients and the thresholds, saved in the decision node and indicated by (2) in the enumeration list of \autoref{sec:ParamsNodes}, got applied to the squared features of the patch to classify it. This process continued until all the validation/test patches of the resolution $r=1$ reached the decision nodes at this resolution layer of the trees and got then decomposed into the voxels by the leaves of the trees. \autoref{fig:RandomForestValidTest} shows the aforementioned steps in processing a validation/test fat-water image.

At the end of this process, each validation/test patch $\rho_{r,j}$ of each resolution $r\in\{0,\cdots,n_{\mathrm{lay}}\}$ visited one decision or leaf node at the bottom of the resolution layer $r$ of every tree in the forest. During each visit, the empirical probabilities, the indicator of classification accuracy, and the indicator of classification difficulty, all saved in the decision/leaf node, got assigned to the patch. Thus, in total, $n_{\mathrm{weak}}=n_{\mathrm{sel}}$ vectors of probabilities ${\big\{\mathbf{p}_{r,w}={[p_{r,w,c}]}_{c\in\mathbb{L}}\big\}}_{w=1}^{n_{\mathrm{weak}}}$, $n_{\mathrm{weak}}=n_{\mathrm{sel}}$ indicators of classification accuracy ${\{(1-\hat{g}^{*}_{r,w})\}}_{w=1}^{n_{\mathrm{weak}}}$, and $n_{\mathrm{weak}}=n_{\mathrm{sel}}$ indicators of classification difficulty ${\{h_{r,w}\}}_{w=1}^{n_{\mathrm{weak}}}$ got assigned to each validation/test patch $\rho_{r,j}$.

From these, a vector of normalized probabilities $\mathbf{a}_{r,j}={\big[a_{r,j,c}\in(0,1)\big]}_{c\in\mathbb{L}}$, a classification label $\hat{l}_{r,j}\in\mathbb{L}$, and an indicator of classification reliability $h^{*}_{r,j}\in(0,1]$ were computed as
\begin{subequations}
\begin{equation}
\label{eq:priorRandForest}
\forall c\in\mathbb{L}:~a_{r,j,c}=\frac{\mathrm{exp}(\overline{a}_{r,j,c})}{\sum_{c'\in\mathbb{L}}\mathrm{exp}(\overline{a}_{r,j,c'})}~~~~~~~~\overline{a}_{r,j,c}=\frac{1}{n_{\mathrm{weak}}}\sum_{w=1}^{n_{\mathrm{weak}}}(1-\hat{g}^{*}_{r,w})\cdot p_{r,w,c}
\end{equation}
\begin{equation}
\label{eq:estClassLabel}
\hat{l}_{r,j}=\argmax_{c}~a_{r,j,c}
\end{equation}
\begin{equation}
\label{eq:diffRandForest}
h^{*}_{r,j}=\mathrm{exp}(-\overline{h}_{r,j})~~~~~~~~~~~~\overline{h}_{r,j}=\frac{1}{n_{\mathrm{weak}}}\sum_{w=1}^{n_{\mathrm{weak}}}h_{r,w}.
\end{equation}
\end{subequations}
That is, first a weighted average of the probabilities was computed with the weight of each decision/leaf node $\eta_{r,w}$ being its classification accuracy $(1-\hat{g}^{*}_{r,w})$. Then, the averaged probabilities $\overline{\mathbf{a}}_{r,j}={\big[\overline{a}_{r,j,c}\big]}_{c\in\mathbb{L}}$ got passed through a Softmax function to obtain the normalized probabilities $\mathbf{a}_{r,j}={\big[a_{r,j,c}\in(0,1)\big]}_{c\in\mathbb{L}}$ with $\sum_{c=1}^{n_{\mathrm{clas}}}a_{r,j,c}=1$. Moreover, the higher the averaged difficulty $\overline{h}_{r,j}$ was, the less reliable the classification of the validation/test patch $\rho_{r,j}$ would be. Thus, we expressed the classification reliability of $\rho_{r,j}$ with $\mathrm{exp}(-\overline{h}_{r,j})$.

If the multiresolution validation/test patches were used to evaluate the proposed forest during/after its optimizations \textbf{alone}, then the estimated labels ${\big\{\hat{\mathbf{l}}_r={[\hat{l}_{r,j}\in\mathbb{L}]}_j\big\}}_{r=0}^{n_{\mathrm{lay}}}$ got evaluated against their corresponding reference labels ${\big\{\mathbf{l}_r={[l_{r,j}\in\mathbb{L}]}_j\big\}}_{r=0}^{n_{\mathrm{lay}}}$. If the multiresolution validation patches were used to evaluate the hyperparameter optimizations of the graphs proposed in \cite{Fallah2017a,Fallah2018p,Fallah2019b,Fallah2021} and the multiresolution test patches were used to evaluate the overall classification performance of the optimized forest followed by the optimized graphs, then the \textbf{multiresolution validation/test samples} needed to be generated from the multiresolution validation/test patches.

Further processing of the multiresolution samples by the multiresolution graphs proposed in \cite{Fallah2017a,Fallah2018p,Fallah2019b,Fallah2021} aimed to improve their classifications. This was due to the fact that the classifications of the forest were irrespective of the spatial (neighborhood) and the hierarchical fine-to-coarse (child-to-parent) relationships of the patches of these samples. The spatial relationships of the patches of each resolution $r\in\{0,\cdots,n_{\mathrm{lay}}\}$ got encoded into a neighborhood graph $\mathcal{G}_r$. The bidirectional hierarchical (parent-to-child and child-to-parent) relationships of the multiresolution patches got encoded into the graph $\mathcal{G}_{\mathrm{hcrf}}$ of a hierarchical conditional random field (HCRF).

The processing of the patches over each graph demanded them to be accompanied with a common set of feature types. That is, for each neighborhood graph $\mathcal{G}_r$ a set of feature types specific to the resolution $r$ and for the graph $\mathcal{G}_{\mathrm{hcrf}}$ a set of feature types valid for all the resolutions were needed. The graphs proposed in \cite{Fallah2017a,Fallah2018p,Fallah2019b,Fallah2021} used the squared features selected by the decision nodes of the trained (optimized) forest. Thus, to obtain a set of feature types (indices) for each resolution $r\in\{0,\cdots,n_{\mathrm{lay}}\}$, indices of the squared features saved in the decision nodes of this resolution of the forest got consolidated. To this end, first the union of these indices was found and then the duplicates in the union got removed. The resulting \textbf{resolution-specific} indices formed a vector of consolidated features denoted by $\tilde{\mathbf{f}}_{r,j}$ for each validation/test patch $\rho_{r,j}$. 

This way, the number of consolidated features for each resolution $r\in\{0,\cdots,n_{\mathrm{lay}}\}$ might differ from other resolutions. This did not matter because each neighborhood graph $\mathcal{G}_r$ procesed its patches independently from other neighborhood graphs in the stack ${\{\mathcal{G}_r\}}_{r=0}^{n_{\mathrm{lay}}}$.

To obtain a set of feature types (indices) which could be used in all the resolutions and thus by the graph $\mathcal{G}_{\mathrm{hcrf}}$, we took the following steps. As listed in \autoref{sec:ParamsNodes}, each decision node also saved the projecting coefficients of its used squared features. The projecting coefficient of each squared feature reflected its importance or relevance for the classifications. Accordingly, we first found the union of all the aforementioned resolution-specific indices collected across all the resolutions $r\in\{0,\cdots,n_{\mathrm{lay}}\}$ and then removed the duplicates. Then, we descendically sorted these consolidated indices according to their projecting coefficients and selected the indices whose coefficients were more than the median of all the coefficients. For a validation/test patch $\rho_{r,j}$, these \textbf{resolution-independent} indices formed a feature vector denoted by $\hat{\mathbf{f}}_{r,j}$.

After the features consolidations, the multiresolution validation/test samples got formed. To this end, each patch $\rho_{r,j}$ of each resolution $r\in\{0,\cdots,n_{\mathrm{lay}}\}$ got accompanied with its
\begin{enumerate}[label={(\arabic*)},leftmargin=*]
\item vector of normalized probabilities $\mathbf{a}_{r,j}={\big[a_{r,j,c}\in(0,1)\big]}_{c\in\mathbb{L}}$ given by \eqref{eq:priorRandForest} and interpreted as its vector of \textbf{prior classification probabilities}
\item indicator of classification reliability $h^{*}_{r,j}\in(0,1]$ given by \eqref{eq:diffRandForest}
\item vector of consolidated resolution-specific features $\tilde{\mathbf{f}}_{r,j}$
\item vector of consolidated resolution-independent features $\hat{\mathbf{f}}_{r,j}$
\item 26 spatial neighbors
\item hierarchical parent in the $(r+1)^{\mathrm{th}}$ resolution layer
\item hierarchical children in the $(r-1)^{\mathrm{th}}$ resolution layer 
\end{enumerate}
to form a validation $v_{r,j}\in\mathbb{D}_{r,\mathrm{val}}$ or a test sample $v_{r,j}\in\mathbb{D}_{r,\mathrm{test}}$ of the resolution $r$.

To optimize the hyperparameters, the multiresolution validation samples and to evaluate the overall classification performance, the multiresolution test samples got processed by the graphs proposed in \cite{Fallah2017a,Fallah2018p,Fallah2019b,Fallah2021}. The processing of the multiresolution samples by the neighborhood graphs ${\{\mathcal{G}_r\}}_{r=0}^{n_{\mathrm{lay}}}$ returned their classification posteriors based on the aforementioned information except for (4), (6), and (7). Then, on the hierarchical graph $\mathcal{G}_{\mathrm{hcrf}}$, these multiresolution posteriors got fused into the multiresolution labels by using the information indicated by (4), (6), and (7) in the above list. The estimated labels got then evaluated against their corresponding reference labels.

\section{Forest's Parameters and Their Optimization}
\label{sec:ParamsForest}
The fixed parameters of the proposed hierarchical quadratic random forest classifier were:
\begin{itemize}[leftmargin=*]
\item number of weak classifiers in the forest $n_{\mathrm{weak}}=n_{\mathrm{sel}}$
\item number of resolution layers of the multiresolution pyramid $n_{\mathrm{lay}}=5$.
\end{itemize}

The tunable parameters of the forest included its main parameters and its hyperparameters.

The main parameters were the projecting coefficients ${[\boldsymbol{\beta}_{r,w,c}\in\mathbb{R}^{n_{\mathrm{feat}}}]}_{c\in\mathbb{L}}$ and the thresholds $\boldsymbol{\tau}_{r,w}\in\mathbb{R}^{n_{\mathrm{clas}}-1}$ of the penalized multivariate linear discriminant of every decision node $\eta_{r,w}$. This discriminant was based on the multiclass sparse discriminant analysis (MSDA) described in \autoref{sec:BackMotivRandForest}. The hyperparameters of the forest and their discretized values were:
\begin{itemize}[leftmargin=*]
\item maximum depth of each tree excluding its leaves: $d_{\mathrm{tree}}=(n_{\mathrm{lay}}-1)+d_1,~d_1\in\{1,\cdots,10\}$
\item maximum Gini impurity of every leaf node: $g_{\mathrm{tree}}\in\{10^{-6},\cdots,10^{-1}\}$
\item regularization parameter of the penalized multivariate linear discriminants of the decision nodes of the resolution layer $r\in\{1,\cdots,n_{\mathrm{lay}}\}$: $\lambda_{r,\mathrm{lda}}\in\{0.01,0.02,\cdots,0.99\}$.
\end{itemize}
This way, $d_1$ and $g_{\mathrm{tree}}$ were the hyperparameters of the resolution layer $r=1$ and $\lambda_{r,\mathrm{lda}}$ was the hyperparameter of every resolution layer $r\in\{1,\cdots,n_{\mathrm{lay}}\}$.

The proposed forest got trained (optimized) to automatically segment $n_{\mathrm{clas}}=|\mathbb{L}|=8$ classes of vertebral bodies (VBs), intervertebral disks (IVDs), psoas major (PM) and quadratus lumborum (QL) muscles, epicardial adipose tissues (EpAT), pericardial adipose tissues (PeAT), cardiac perivascular adipose tissues (PvAT), and background on each volumetric fat-water image. The main parameters and the hyperparameters of the forest got optimized from the coarsest resolution to the finest resolution. To this end, each resolution layer $r\in\{1,\cdots,n_{\mathrm{lay}}\}$ got built over the previously built (coarser) resolution layers by conducting several optimization trials for it. In each trial, a set of hyperparameter values got randomly selected. With these hyperparameter values, the main parameters of the current resolution layer got optimized by using the multiresolution training samples according to the process described in \autoref{sec:TrainRF}. The classification performance of the resulting forest got then evaluated on the multiresolution validation patches according to the process described in \autoref{sec:MultResValTestSmpls}. These evaluations were done by calculating the \textit{precision} and the \textit{recall} metrics of the estimated classification labels $\hat{\mathbf{l}}_r={[\hat{l}_{r,j}\in\mathbb{L}]}_j$ against their corresponding reference labels $\mathbf{l}_r={[l_{r,j}\in\mathbb{L}]}_j$ for each of the $n_{\mathrm{clas}}-1=8-1=7$ foreground classes against the rest of the classes. This way, 7 \textit{precision} and 7 \textit{recall} values were obtained. These values got averaged to represent the overall classification performance of the forest resulted from the selected hyperparameter values and the main parameters optimized with them. The optimization trials continued by randomly selecting another set of hyperparameter values until the resulting forest could not exceed the averaged \textit{precision} and \textit{recall} values of any of the forests in the last 20 trials.

The \textit{precision} and \textit{recall} metrics were selected due to their robustness against the imbalanced class-sample distributions. We drew the idea of maximizing the averaged (across-class) \textit{precision} and \textit{recall} metrics from the procedure of optimizing a linear discriminant addressing a binary classification. In this procedure, the optimum parameters of the linear discriminant were the maximizers of the area under the \textit{precision-recall curve}. Each \textit{precision-recall curve} got obtained by varying the decision threshold between the projected features of the addressed class and the projected features of the rest of the classes and computing the \textit{precision} and the \textit{recall} metrics for each value of the decision threshold \cite{Brownlee2020}. In our case, the main parameters including the projecting coefficients and the thresholds got optimized according to the approach proposed in \autoref{sec:TrainRF}. Thus, we only maximized the averaged (across-class) \textit{precision} and \textit{recall} metrics with regard to the selected hyperparameter values.

\autoref{table:hypOptRandForest} shows the optimized hyperparameters and the overall time of feature extraction and optimization of the main parameters and the hyperparameters of the proposed forest on a PC with 16 GB RAM and a quad-core CPU of 3.10 GHz frequency. The independence of the decision trees of the forest due to the chosen bagging-based ensemble collection allowed the trees to be optimized in parallel. This could speed up the optimizations in comparison to a sequential booting-based ensemble collection described in \autoref{sec:BackMotivRandForest}.

The increase of the regularization parameter $\lambda_{r,\mathrm{lda}}$ from the resolution $r=5$ to the resolution $r=3$ could be attributed to an increased need to the penalization due to an increased classification variance (overfitting). This increase was simultaneous with a decrease of the classification bias due to the smaller samples resulted from the hierarchical decomposition of the samples from the resolution $r=5$ to the resolution $r=3$. However, by further decomposing the samples from the resolution $r=3$ to the resolution $r=1$, the number of the samples increased so much that it could partially mitigate the overfitting. This in turn reduced the need to the penalization and thus reduced the penalization parameter $\lambda_{r,\mathrm{lda}}$ in moving from the resolution $r=3$ to the resolution $r=1$.

\begin{table}[t!]
\begin{center}
\caption{Optimized hyperparameters and the time of building and parallelly optimizing the decision trees of the proposed forest with 50 training volumetric fat-water images.}
\label{table:hypOptRandForest}
\resizebox{1.0\textwidth}{!}{%
\begin{tabular}{|ccccccc|}
\hline
\multicolumn{7}{|c|}{\textbf{Hyperparameters}}\\
$d_1$&$g_{\mathrm{tree}}$&$\lambda_5$&$\lambda_4$&$\lambda_3$&$\lambda_2$&$\lambda_1$\\\hline
4&$10^{-3}$&0.15&0.38&0.52&0.27&0.18\\\hline
\end{tabular}
\begin{tabular}{|cccc|}
\hline
\multicolumn{4}{|c|}{\textbf{Time}}\\
\text{Building}&\text{Feature}&\text{SMOTE}&\text{Trees}\\
\text{Pyramid}&\text{Extraction}&\text{Sampling}&\text{Optimization}\\\hline
2.7 min&2.5 h&15 min&38 min\\\hline
\end{tabular}}
\end{center}
\end{table}

\begin{singlespace}
{\footnotesize
\bibliography{ArtklBookMisc,IEEEabrvIndexMedicus,ProcLong}}
\end{singlespace}
\end{document}